\documentclass[10pt, a4paper]{article}

\usepackage{lrec-coling2024} 
\usepackage{amsfonts, amsmath, amssymb}
\usepackage{booktabs}
\usepackage{hyperref}
\usepackage{multirow, multicol, makecell}
\newcommand{\nop}[1]{}
\usepackage{spverbatim}
\usepackage{fancyvrb}
\usepackage{bbm}
\usepackage{subfigure}
\usepackage{tikz, pgfplots}
\usepackage{marvosym}

\definecolor{mygreen}{RGB}{112,173,71}
\definecolor{myblue}{RGB}{46,117,182}
\definecolor{myred}{RGB}{255,0,0}

\title{\textbf{ToNER: Type-oriented Named Entity Recognition \\with Generative Language Model}}

\name{\begin{tabular}{c}
     Guochao Jiang$^\dag$, Ziqin Luo$^\dag$, Yuchen Shi$^\dag$, Dixuan Wang$^\dag$, \\
     Jiaqing Liang$^{\dag\ddag}$, Deqing Yang$^{\dag\ddag\textrm{\Letter}}$
\end{tabular}}

\address{$^\dag$School of Data Science, Fudan University\\
         $^\ddag$Shanghai Key Laboratory of Data Science\\
         $^\dag$\texttt{\{gcjiang22, zqluo22, ycshi21, dxwang23\}@m.fudan.edu.cn}\\
         $^\ddag$\texttt{\{liangjiaqing, yangdeqing\}@fudan.edu.cn}\\}

\abstract{
In recent years, the fine-tuned generative models have been proven more powerful than the previous tagging-based or span-based models on named entity recognition (NER) task. It has also been found that the information related to entities, such as entity types, can prompt a model to achieve NER better. However, it is not easy to determine the entity types indeed existing in the given sentence in advance, and inputting too many potential entity types would distract the model inevitably. To exploit entity types' merit on promoting NER task, in this paper we propose a novel NER framework, namely \emph{ToNER} based on a generative model. In ToNER, a type matching model is proposed at first to identify the entity types most likely to appear in the sentence. Then, we append a multiple binary classification task to fine-tune the generative model’s encoder, so as to generate the refined representation of the input sentence. Moreover, we add an auxiliary task for the model to discover the entity types which further fine-tunes the model to output more accurate results. Our extensive experiments on some NER benchmarks verify the effectiveness of our proposed strategies in ToNER that are oriented towards entity types' exploitation.\footnotemark
 \\ \newline \Keywords{Named Entity Recognition, Natural Language Generation, Information Extraction, Information Retrieval
 } 
 }

\begin{document}

\maketitleabstract
\footnotetext[1]{Our code is available at \url{https://github.com/jiangguochaoGG/ToNER}.}
\section{Introduction}\label{sec:intro}
As one representative task of information extraction, \emph{named entity recognition} (NER) \cite{NER} has been the critical component to achieve plenty of downstream applications, such as the construction of knowledge graph \cite{CNDB}, information retrieval \cite{banerjee2019information}, question answering \cite{molla2006named}, and recommendation \cite{madani2022review}. The primary objective of NER is to identify the span(s) of entity mention(s) within a given input sentence and subsequently categorize each identified entity.

\begin{figure}[t]
    \centering
    \includegraphics[width=0.45\textwidth]{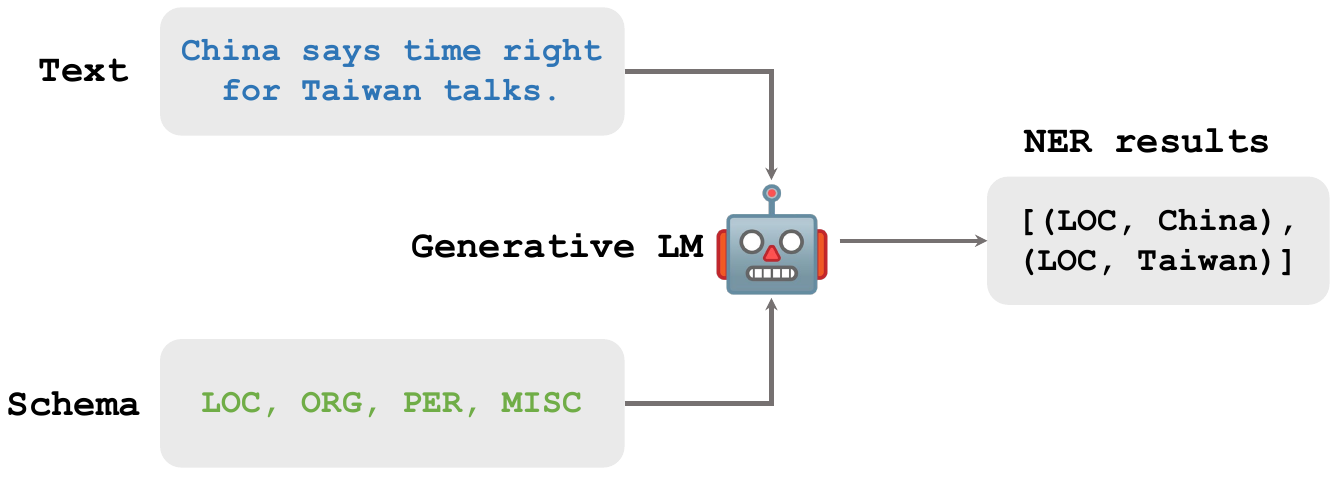}
    \caption{The standard inputs and outputs for a generative model to achieve NER. Besides the given sentence, all candidate entity types regarded as the schema are also input into the model as the prompt.}
    \label{fig:schema}
    \vspace{-0.2cm}
\end{figure}

The previous NER solutions include tagging-based models \cite{strubell-etal-2017-fast, BERT, conllpp, sharma2019bioflair, 10.1093/bioinformatics/btz682} and span-based models \cite{yu-etal-2020-named, li-etal-2020-unified, li2022unified, wang-etal-2022-miner, fu-etal-2021-spanner, li-etal-2021-span}. In recent years, some researchers have employed the generative pre-trained language models such as T5 \cite{T5}, BART \cite{BART} and GPT-3 \cite{GPT-3} to achieve NER, given their powerful capability of natural language generation. According to the input requirement of generative models, the given sentence and the candidate entity types are simultaneously input into the model as the prompt to trigger the generation of NER results. As shown in Figure \ref{fig:schema}, besides the sentence ``China says time right for Taiwan talks.'', the candidate entity type LOC, ORG, PER and MISC regarded as the schema, are also input into the model as the prompt. Then, the model would generate the entity span `China' and `Taiwan' existing in the sentence, and simultaneously assign the correct type LOC from the schema for each of them.

It has been found that leveraging the information of entity types can help the model recognize the entities in the sentence more accurately \cite{10094905, li2023type}. In general, there are many entity types in one NER corpus. It inevitably increases the difficulty of achieving accurate NER if too many types are input as the model's prompt. 
In addition, it is non-trivial to infer the entity types that are more likely to appear in the sentence in advance, and by now there are still no effective methods of discovering such entity types that can be directly applied to generative models.

To address these problems, we propose a novel NER framework, namely \emph{ToNER} (\textbf{T}ype-\textbf{o}riented \textbf{N}amed \textbf{E}ntity \textbf{R}cogition), which takes a generative model as the backbone and fully leverages the entity types to achieve enhanced NER. Specifically, we first introduce a small model to compute the matching degree between each candidate entity type and the input sentence, which is used to identify the types mostly likely to appear in the sentence. It helps the generative model concentrate on a limited number of credible types during achieving NER task. In addition, we add an additional multiple binary classification task to fine-tune the encoder in the generative model, so as to obtain optimal sentence representation which is benefit to generate more accurate NER results. Inspired by \cite{wang2022instructionner, lu2022unified, wang2023instructuie}, we further propose an auxiliary task for the generative model to recognize all entity types in the input sentence, which is different to the primary NER task but further fine-tunes the model to generate more accurate NER results.

Our main contributions in this paper are summarized as follows.

1. We propose a novel NER framework \emph{ToNER} which successfully combines a generative pre-trained language model with a relatively small matching model to achieve more accurate NER.

2. We not only introduce a type matching model to discover the entity types most likely to appear in the input sentence, but also propose auxiliary learning tasks for fine-tuning the generative model, all of which can help ToNER obtain improved NER performance. 

3. 
Our extensive experiments demonstrate that the proposed ToNER almost achieves the state-of-the-art (SOTA) performance on multiple representative NER benchmarks, and the effectiveness of each component in ToNER we propose is also justified, including the impact of adding Chain-of-Thoughta(CoT)-style explanations.
\section{Related Work}

\paragraph{Named Entity Recognition}
The task of Named Entity Recognition (NER) aims to identify spans expressing entities from text \cite{conll2003}, including three tasks: flat NER, nested NER, and discontinuous NER. Nested NER includes overlapping entities, while entities in discontinuous NER may include multiple nonadjacent spans. Overall, NER models can be divided into three types: those based on token sequence labeling, span classification, and seq2seq generation. In token-level models \cite{ratinov2009design,strakova2019neural,dai2020effective}, each token is tagged as BIO or BILOU, and decoded using Conditional Random Fields (CRF) or other methods. In the category of span-level classification methods \cite{wang2020pyramid,Yu2020NamedER}, the text within the span is considered as a whole and classified using a classification model to determine whether it is an entity. Some methods based on hypergraphs \cite{lu2015joint,Wang2018NeuralSH} also fall into this category. In seq2seq generative models, the extraction target is encoded as a text sequence. Various methods \cite{cui2021template,yan2021unified} have explored different forms of input text and target coding, which we will introduce more in the next section.

\paragraph{Generative Methods for NER}
Benefiting from the development of generative Pretrained Language Models (PLMs), more and more work has adopted a sequence-to-sequence (seq2seq) approach to complete NER tasks. \cite{cui2021template} models the NER task as a template filling task, using PLMs to fill in candidate spans and entity categories in pre-written templates. However, enumerating all possible spans is time-consuming. \cite{yan2021unified} transforms flat NER, nested NER, and discontinuous NER into a unified entity span sequence generation problem, and proposes a pointer-based framework based on BART to infer entity boundaries and categories simultaneously. Building on this, \cite{chen2021lightner} introduces prompt-tuning to the attention mechanism of BART for low-resource scenarios. \cite{wang2022instructionner} introduces task instructions and answer options in the input sentence, and directly extracts the required entity as the target output, by instructing the tuning of the T5 model. Compared to these works, our model adopts, and emphasizes, the role of entity type, and designs targeted auxiliary tasks.

\begin{figure*}[t]
    \centering
    \includegraphics[width=0.95\textwidth]{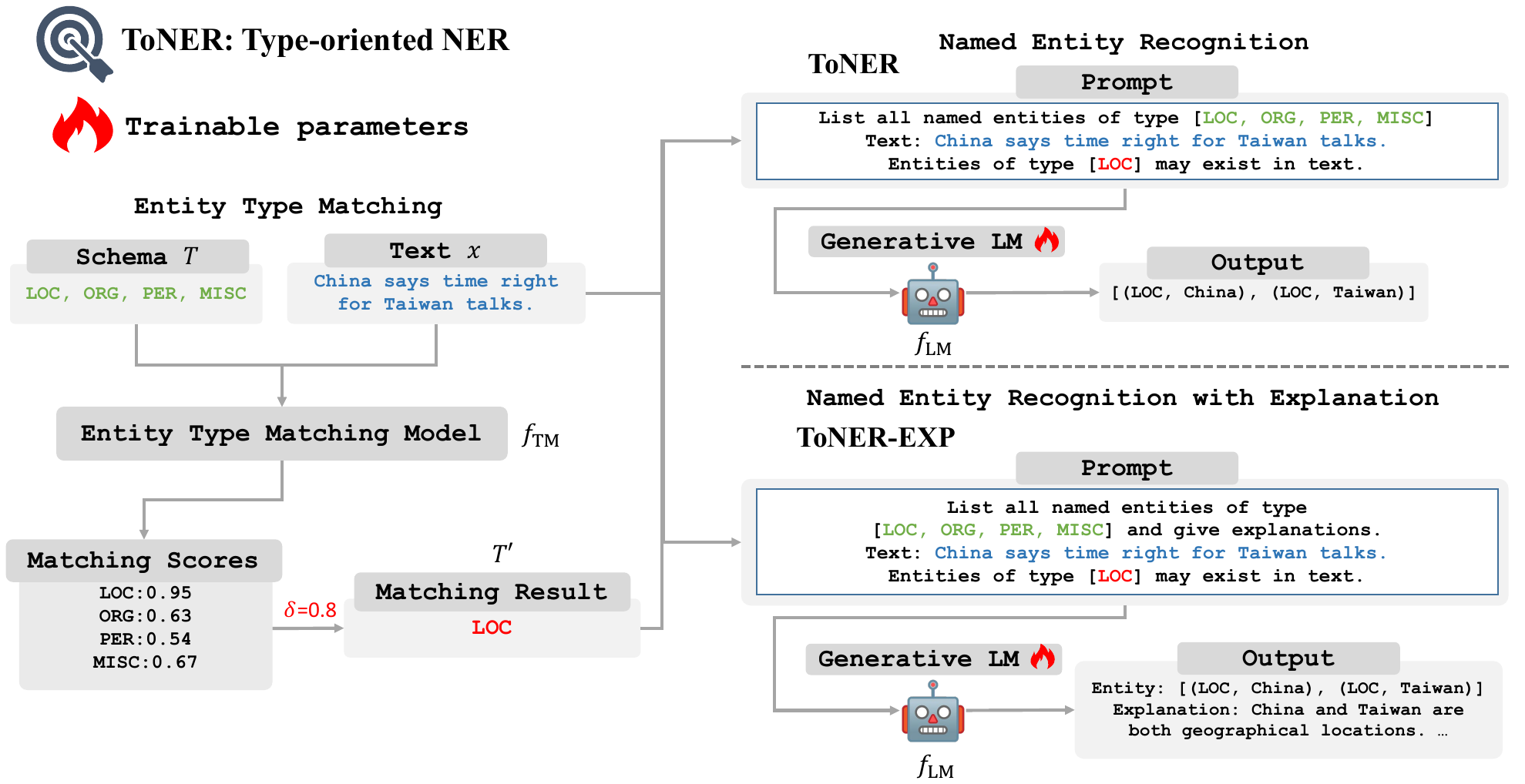}
    \caption{The pipeline of our proposed ToNER. ToNER achieves the NER task mainly with a generative model $f_\text{LM}$, which focuses more on the entity type LOC filtered out by the matching model $f_\text{TM}$. In the figure, the schema part is represented by \textcolor{mygreen}{green} characters, the text to be extracted is represented by \textcolor{myblue}{blue} characters, and the entity type matching results are represented by \textcolor{myred}{red} characters.}
    \vspace{-0.2cm}
    \label{fig:ToNER}
\end{figure*}

\paragraph{Multi-task Learning in Information Extraction}
Many previous works \cite{wang2022instructionner,wang2023instructuie} have validated that introducing relevant intermediate tasks or auxiliary tasks in information extraction tasks can enhance the overall performance of the model. \cite{lu2022unified} models IE as a unified text-to-structure task. Besides the main extraction task, Universal Information Extraction (UIE)\cite{lu2022unified} also introduces an auxiliary learning task for the intermediate structured extraction language. \cite{wang2022instructionner}, besides the main NER task, introduces two auxiliary tasks - entity extraction and entity typing - which respectively enhance the model's ability to capture entity boundaries, and understand entity category information. Through ablation experiments, these two auxiliary tasks have been found to improve NER performance, especially in low-resource NER settings. \cite{wang2023instructuie} also validates that auxiliary tasks can provide additional information that complements the main task. For Named Entity Recognition, relationship extraction, and event extraction tasks, they designed span extraction and entity typing tasks, an entity pair extraction task and a relationship classification task, and a trigger extraction task and an argument extraction task, respectively.
\section{Methodology}
In this section, we first introduce the basic framework of achieving NER with a generative model, and then present our special designs in our ToNER to obtain enhanced NER performance.

\begin{table*}[t]
\centering
\resizebox{\textwidth}{!}{
\begin{tabular}{c|l}
\toprule
$t_i$ & $D_{t_i}$ \\
\bottomrule
LOC & location: Names that are locations. \\
PER & person: Names of people. \\
ORG & organization: Companies, agencies, institutions, etc. \\
MISC & miscellaneous: Names of miscellaneous entities that do not belong to person, organization and location. \\
\bottomrule
\end{tabular}
}
\vspace{-0.2cm}
\caption{The descriptions for some representative entity types in CoNLL2003 dataset. These descriptions provide the matching model with richer semantic information of entity types.}
\label{table:matching_text}
\vspace{-0.2cm}
\end{table*}

\subsection{Named Entity Recognition with A Generative Model}
Formally, we denote the generative language model as $f_\text{LM}$, the input token sequence as $x = \{x_1, x_2, \cdots, x_m\}$, and the input instruction as $\mathcal{I}$. In addition, the output (generated) token sequence is denoted as $y = f_\text{LM}(x) =\{y_1, y_2, \cdots, y_n\}$. 
For the classic auto-regressive generative model, the sampling probability of the model generating $y$ is formularized as

\begin{align}
    \mathbb{P}(y | \mathcal{I}, x) = \prod_{t=1}^n \mathbb{P} (y_t | \mathcal{I}, x, y_{<t}).
\end{align}

In our ToNER, we input the following prompt into the generative model to achieve NER,

\noindent\texttt{List all named entities of type [$T$]}

\noindent\texttt{Text: $x$}

\noindent Wherein $T$ is the list of candidate entity types, i.e., the input schema.

Using generative models to achieve information extraction generally requires the model to output the results according to a given format. 
In ToNER, the generative model's outputs follow the format as

\noindent\texttt{[(type$_1$, entity$_1$), (type$_2$, entity$_2$), ..., (type$_l$, entity$_l$)]}

\noindent Among them, \texttt{type$_i(1\leq i\leq l)$} is the type assigned to the extracted (generated) entity span \texttt{entity$_i$}. 

\nop{
For example, taking a sample from the CoNLL2003 dataset \cite{conll2003}.
\begin{Verbatim}[commandchars=\\\{\}]
\textbf{Input} \quad List all named entities of the type 
[location, person, organization, miscellaneous]
Text: BayerVB sets C\$ 100 million six-year bond .
\textbf{Output} \quad [(organization, BayerVB), 
(miscellaneous, C\$)]
\end{Verbatim}
}

According to the generative model's rule of generating tokens, the loss of generating $y$ is as follows,
\begin{align}
    \mathcal{L}_g = -\sum_{t=1}^n \log \mathbb{P} (y_t | \mathcal{I}, x, y_{<t}).
\end{align}

\subsection{Entity Type Matching Model}
\label{sec:tm}

As we introduced before, a predefined (candidate) list of entity types should be input as the schema into the generative model, to trigger the generation of NER. With such a prompt, the model needs to fully understand the semantics of each given entity type, based on which it then assigns the correct type for each generated entity span. 
This procedure implies that, too many candidate entity types would hinder the model from assigning the correct types for the entities in the sentence. As a result, reducing the entity types deserving to be cared about is the key to enhance the model's NER performance.


To this end, we introduce an entity type matching model in our ToNER, denoted as $f_\text{TM}$, which computes the semantic similarity, i.e., the matching degree between each type and the sentence based on their semantic representation.
Thus, the entity types most likely to appear in the sentence can be identified to reduce the number of entity types on which the model should concentrate. 

Formally, suppose the original candidate entity type list (schema) is $T=\{t_1, t_2, \cdots, t_k\}$. Given that information of encoding $t_i (1\leq i\leq k)$ is not sufficient to compute the accurate matching degree between $t_i$ and $x$, we incorporate an additional description for $t_i$, of which the token sequence is denoted as $D_{t_i}$. Obviously, $D_{t_i}$ contains more richer semantic information of $t_i$. Taking the dataset CoNLL2003 as an example, we list the descriptions for some representative entity types in Table \ref{table:matching_text}, which were provided in the original paper \cite{conll2003} .

Specifically, we adopt a BERT-like architecture for Entity Type Matching Model's encoder, denoted as $E_\text{TM}$, which converts a piece of input text into a representation through average pooling the last hidden state of each token in the text. Given $x$ and a candidate $t$, the whole entity type matching model $f_\text{TM}$ outputs the semantic similarity between $x$ and $t$ as
\begin{align}\label{eq:score}
    f_\text{TM}(x, t) = \frac{E_\text{TM}(x)^\top E_\text{TM}(D_{t})}{\| E_\text{TM}(x) \|_2 \| E_\text{TM}(D_{t}) \|_2},
\end{align}
where $E_\text{TM}(x)\in\mathbb{R}^d$ is $x$'s semantic representation generated by the encoder. 
Thus, the entity types in $T$ with the semantic similarity score higher than the threshold $\delta$ are retained as the possible types in the sentence, which constitute a new schema denoted as $T'$.

Next, with $T'$ we modify the prompt of the generative model as follows, 

\noindent\texttt{List all named entities of type [$T$]}

\noindent\texttt{Text: $x$}

\noindent\texttt{Entities of type [$T'$] may exist in text.}

\noindent With such a prompt, the generative model can focus more on the types in $T'$ rather than $T$, which reduces the difficulty of achieving NER with the model. We still list the original schema $T$ in the prompt to ensure the model does not miss the correct entity types not in $T'$.

The pipeline of incorporating the entity types filtered out based on the matching model into the generative model is shown in Figure \ref{fig:ToNER}.

In order to train $f_\text{TM}$, we have collected sufficient samples from the original NER benchmark. Formally, suppose the sets of entity types that are mentioned and not mentioned in $x$ are denoted as $\mathcal{P}_x$ and $\mathcal{N}_x$, respectively, then inspired by SimCSE \cite{SimCSE}, we propose the following the loss to train $f_\text{TM}$,
\begin{align}
    \mathcal{L}_m = - \sum_{t^+ \in \mathcal{P}_x} \log \frac{\mathrm{e}^{f_\text{TM}(x, t^+) / \tau}}{\sum_{t \in \mathcal{P}_x \bigcup \mathcal{N}_x}\mathrm{e}^{f_\text{TM}(x, t) / \tau}},
\label{loss:mathcing_model}
\end{align}
where $\tau$ is a hyperparameter of temperature .

\subsection{Fine-tuning Encoder with Type Classification}\label{sec:tc}
For a model with an encoder-decoder architecture, the encoder is its critical component since the model's results are generated mainly based on the representations learned by the encoder. As we know, the generative pre-trained language models are pre-trained through the task different to NER, although they can directly achieve NER task. Thus, we believe that fine-tuning the encoder in ToNER with the task more related to NER would help the encoder generate refined representations in terms of improved NER. Since we have found that the entity types existing in the sentence are helpful, we propose a multiple binary classification task as an auxiliary learning task of ToNER to train a better encoder, resulting in more accurate generations of NER.


Formally, suppose $h(x)$ is $x$'s representation which is generated by the encoder through the average pooling upon the hidden states of all tokens in $x$. Then, we adopt a neural classifier $c$ to map $h(x)$ to a $k$-dimensional vector as $c\big(h(x)\big) = [p_1, p_2, \cdots, p_k] \in \mathbb{R}^k$, where $p_i (1\leq i\leq k)$ is the logit corresponding to the candidate entity types $t_i$, indicating whether $t_i$ appears in $x$ or not. 
The loss of this multiple binary classification is as follows,
\begin{equation}
\begin{split}
    \mathcal{L}_c &= \log \left( 1+ \sum_{i = 1}^k \mathbbm{1} \left\{ t_i \in \mathcal{P}_x \right\} \mathrm{e}^{p_i} \right) + \\ &\log \left( 1+ \sum_{i = 1}^k \mathbbm{1} \left\{ t_i \in \mathcal{N}_x \right\} \mathrm{e}^{-p_i} \right),
\end{split}
\end{equation}
where $\mathbbm{1}\{ \}$ is indicator function.

Thus, the overall training loss for ToNER is 
\begin{align}
    \mathcal{L} = \mathcal{L}_g + \lambda \mathcal{L}_c,
    \label{loss:all}
\end{align}
where $\lambda$ is the controlling parameter.

\nop{
\begin{figure*}[t]
    \centering
    \includegraphics[width=0.8\textwidth]{Figures/encoder-decoder.pdf}
    \caption{The overview of ToNER's training procedure.}
    \label{fig:encoder-decoder}
\end{figure*}
}

\subsection{Improving NER by An Auxiliary Task}\label{sec:tr}
Many previous works \cite{wang2022instructionner, lu2022unified, wang2023instructuie} have verified that for information extraction tasks, adding relevant intermediate tasks or auxiliary tasks (such as entity typing, entity extraction and relation extraction) in the instruction fine-tuning stage is beneficial to improve the model's performance on the primary extraction task. Inspired by them, we also add an auxiliary task in the instruction fine-tuning stage to explicitly encourage the model recognize the entity types that may exist in the input sentence.

Similar to the instruction prompt of NER, we construct the following prompt to ask the generative model to list all entity types in the sentence. 

\noindent\texttt{List all entity types in the text from type [$T$]}

\noindent\texttt{Text: $x$}

To construct this auxiliary task's training samples, we randomly select some training samples from the datasets, each of which only takes the entity types as its label (model output). Obviously, if the generative model can accomplish this auxiliary task well, it can also generate satisfactory NER results since these two tasks are very correlated.

\subsection{Achieving NER with CoT-style Explanations}
\emph{Chain-of-Thought} (CoT) Prompting \cite{DBLP:journals/corr/abs-2201-11903} has been widely used to improve the performance of large language models (LLMs) on various NLP tasks. A recent study \cite{wadhwa-etal-2023-revisiting} has found that using CoT-style explanations generated by LLMs to fine-tune the relatively small generative models can help improve the small models' performance on relation extraction. It inspires us to investigate whether adding CoT-style explanation besides outputting the standard NER results could improve our ToNER's performance.

To this end, for each training sample in the datasets we used an LLM (such as ChatGPT\footnote[2]{\url{https://openai.com/blog/chatgpt/}} or GPT4 \cite{openai2023gpt4}) to generate the explanations for its NER result. Next, we used these augmented samples with CoT-style explanations to fine-tune the generative model, and thus get a advanced variant of ToNER, denoted as ToNER-EXP. In the following experiments, we will compare the performance of ToNER and ToNER-EXP. We list two instances of such training samples from CoNLL2003 dataset as follows.

\noindent\texttt{\textbf{Input} List all named entities of type [organization, person, miscellaneous, location] and give explanations. 
}

\noindent\texttt{Text: -- Wellington newsroom 64 4 4734 746
}

\noindent\texttt{Entities of type [location] may exist in text.}

\noindent\texttt{\textbf{Output} Entity: [(location, Wellington)]
}

\noindent\texttt{Explanation: 'Wellington' is labeled as 'location' because it refers to a specific location, which is the capital city of New Zealand.
}

\noindent====================================

\noindent\texttt{\textbf{Input} List all named entities of type [organization, person, miscellaneous, location] and give explanations.
}

\noindent\texttt{Text: The bank said there were concerns fiscal consolidation would unduly restrict growth, but evidence was ambiguous.
}

\noindent\texttt{Entities of type [] may exist in text.}

\noindent\texttt{\textbf{Output} Entity: []}

\noindent\texttt{Explanation: No entity in the text belongs to any pre-defined entity type.
}


\section{Experiment}
In this section, we evaluate the performance of our ToNER with some previous NER models on several NER datasets, and further analyze the experiment results.

\begin{table*}[t]
\centering
\resizebox{\textwidth}{!}{
\begin{tabular}{lll}
\toprule
Para. & Value & Comment \\\hline
batch size & 32 or 8 & 32 for Flan-T5-large and Flan-T5-xl, 8 for Flan-T5-xxl \\
max length & 512 & max token length for input and output \\
$lr$ & 3e-5 & learning rate for ToNER \\
$\lambda$ & 0.1 & the controlling parameter of $\mathcal{L}_{c}$ in Eq. \ref{loss:all} \\
$\tau$ & 0.05 & temperature coefficient for matching model loss \ref{loss:mathcing_model} \\
$\delta$ & 0.8/0.7/0.6 & type matching threshold of CoNLL2003/JNLPBA/OntoNotes 5.0, ACE2004, ACE2005\\
\bottomrule
\end{tabular}
}
\vspace{-0.3cm}
\caption{Some important hyperparameter settings for ToNER implementation.}
\label{tb:hyper_parameters}
\end{table*}

\begin{table*}[t]
\centering
\resizebox{0.8\textwidth}{!}{
\begin{tabular}{c|ccc|ccc} 
\Xhline{1.2pt} 
 \multirow{2}*{\textbf{Model}} &  \multicolumn{3}{c|}{\textbf{CoNLL2003}} & \multicolumn{3}{c}{\textbf{OntoNotes 5.0}} \\
 & {P} & {R} & {F1} & {P} & {R} & {F1} \\ 
 \Xhline{1pt} 
\citeauthor{strubell-etal-2017-fast} (\citeyear{strubell-etal-2017-fast}) & - & - & 90.65  & - & - & 86.84 \\
\citeauthor{BERT} (\citeyear{BERT}) & - & - & 92.80 & 90.01 & 88.35 & 89.16 \\
\citeauthor{yu-etal-2020-named} (\citeyear{yu-etal-2020-named}) & 92.91 & 92.13 & 92.52 & 90.01 & 89.77 & 89.89 \\ 
\citeauthor{li-etal-2020-unified} (\citeyear{li-etal-2020-unified}) & 92.33 & \textbf{94.61} & 93.04 & \textbf{92.98} & 89.95 & 91.11 \\
\citeauthor{yan2021unified} (\citeyear{yan2021unified}) & 92.61 & 93.87 & 93.24 & 89.99 & 90.77 & 90.38 \\
\citeauthor{li2022unified} (\citeyear{li2022unified}) & 92.71 & 93.44 & 93.07 & 90.03 & 90.97 & 90.50 \\
\citeauthor{wang2023instructuie} (\citeyear{wang2023instructuie}) & - & - & 92.94 & - & - & 90.19 \\ \citeauthor{shen-etal-2023-diffusionner} (\citeyear{shen-etal-2023-diffusionner}) & 92.99 & 92.56 & 92.78 & 90.31 & 91.02 & 90.66 \\ \Xhline{1pt}
$\mathrm{ToNER_{large}}$ & \textbf{93.55} & \underline{94.11} & \textbf{93.83} & \underline{91.16} & \underline{91.35} & \underline{91.25} \\
$\mathrm{ToNER_{xl}}$ & \underline{93.53} & 93.65 & \underline{93.59} & 91.11 & \textbf{91.50} & \textbf{91.30} \\
\Xhline{1.2pt}
\end{tabular}
}
\vspace{-0.2cm}
\caption{All models' NER performance on CoNLL2003 and OntoNotes 5.0. \textbf{Bold} and \underline{underline} denote the best and second best scores, respectively.
}
\label{table:CoNLL2003}
\end{table*}

\subsection{Datasets}
We conducted our experiments on the following five NER benchmarks.

\noindent 1. \textbf{CoNLL2003} \citeplanguageresource{conll2003_lr} is a collection of news wire articles from the Reuters Corpus, which contains 4 entity types including \verb|LOC|, \verb|ORG|, \verb|PER| and \verb|MISC|.

\noindent 2. \textbf{OntoNotes 5.0} \citeplanguageresource{ontonotes_lr} is a large corpus comprising various genres of text (news, conversational telephone speech, weblogs, usenet newsgroups, broadcast, talk shows). We only considered its English samples in our experiments.

\noindent 3. \textbf{JNLPBA} \citeplanguageresource{JNLPBA_lr} is a biomedical dataset from the GENIA version 3.02, which contains 5 entity types including \verb|DNA|, \verb|RNA|, \verb|cell_type|, \verb|cell_line| and \verb|protein|. It was created with a controlled search on MEDLINE.

\noindent 4. \textbf{ACE2004} \citeplanguageresource{ace2004_lr} and \textbf{ACE2005} \citeplanguageresource{ace2005_lr} are two nested NER datasets, which contain 7 entity types including \verb|PER|, \verb|ORG|, \verb|LOC|, \verb|GPE|, \verb|WEA|, \verb|FAC| and \verb|VEH|, and are generally used to evaluate the more complicate task of overlapped NER. We followed the same setup as the previous work \cite{katiyar-cardie-2018-nested, lin-etal-2019-sequence}.

\subsection{Baselines}
We compared our ToNER with the following NER baselines belonging to different families, including the tagging-based methods \cite{strubell-etal-2017-fast, BERT, conllpp, sharma2019bioflair, 10.1093/bioinformatics/btz682}, the span-based methods \cite{yu-etal-2020-named, li-etal-2020-unified, li2022unified, wang-etal-2022-miner, fu-etal-2021-spanner, li-etal-2021-span}, and the generation-based methods \quad \cite{yan2021unified, wang2023instructuie, shen-etal-2023-diffusionner, Fei_Ji_Li_Liu_Ren_Li_2021}.

\subsection{Implementation Details}
We report the entity-level micro Precision (P), Recall (R) and F1 scores of all compared models in the following result figures and tables. To construct ToNER, we selected Flan-T5 \cite{Flan-T5} as the generative model in our framework. We used AdamW \cite{adamw} as our models' optimizer. We also selected GTE-large \cite{gte} as the entity type matching model and fine-tuned it with the AdamW of 1 epoch, where the learning rate is 8e-6, and the weight decay is 1e-3. The CoT-style explanations were generated by GPT4. Other important hyperparameters are listed in Table \ref{tb:hyper_parameters}, which were decided based on our tuning studies. We conducted our experiments on eight NVIDIA Tesla A100 GPU with 80GB of GPU memory.

\subsection{Overall Performance Comparisons}
For the baselines' NER performance on the different datasets, we directly report their results published on the previous paper. Thus different baselines are reported in the result  Table \ref{table:CoNLL2003} $\sim$ \ref{table:ACE2004} corresponding to different datasets, where the best scores and the second best scores are bold and underlined, respectively. 
Specifically, we compared our ToNER with Flan-T5-large and Flan-T5-xl with the baselines, which are denoted as $\mathrm{ToNER_{large}}$ and $\mathrm{ToNER_{xl}}$, respectively.

From the tables, we find that on CoNLL2003, OntoNotes 5.0 and JNLPBA, our $\mathrm{ToNER_{large}}$ or $\mathrm{ToNER_{xl}}$ achieves the best F1. Since CoNLL2003 and OntoNotes 5.0 are both NER datasets in general fields, and  JNLPBA is the dataset of the biological field, it shows that our ToNER is more effective than the baselines in both general fields and the special field.  
We also compared our framework's performance with the baselines on ACE2004 and ACE2005 for the task of overlapped NER, of which the results are listed in Table \ref{table:ACE2004}. Although the baseline \citeauthor{shen-etal-2023-diffusionner} (\citeyear{shen-etal-2023-diffusionner}) achieves the best F1 on this task, our $\mathrm{ToNER_{xl}}$ can also obtain quite competitive performance. 
\nop{For CADEC, ToNER achieves 69.68\% F1 using Flan-T5-large. ToNER relies on the entity type matching model to a certain extent. Intuitively, the more entity types involved in a task, the more information the matching model can provide to the base model. When the task involves few entity types, the effect improvement brought by the matching model is limited. CADEC dataset only has 1 entity type \verb|ADEs|. Too few entity types in the task limit the information gain of the matching model, which may be the reason why our method is inferior to other existing models.}

\begin{table}[t]
\centering
\resizebox{0.5\textwidth}{!}{
\begin{tabular}{c|ccc} 
\Xhline{1.2pt} 
 \multirow{2}*{\textbf{Model}} &  \multicolumn{3}{c}{\textbf{JNLPBA}}  \\
 & {P} & {R} & {F1}\\ 
 \Xhline{1pt} 
 \citeauthor{fu-etal-2021-spanner} (\citeyear{fu-etal-2021-spanner}) & - & - & 74.49 \\
 \citeauthor{wang-etal-2022-miner} (\citeyear{wang-etal-2022-miner}) & - & - & 77.03 \\ \citeauthor{sharma2019bioflair} (\citeyear{sharma2019bioflair}) & - & - & 77.03 \\
 \citeauthor{10.1093/bioinformatics/btz682} (\citeyear{10.1093/bioinformatics/btz682}) & 72.68 & \textbf{83.21} & 77.59 \\
 \Xhline{1pt}
$\mathrm{ToNER_{large}}$ & \textbf{77.88} & 79.55 & \underline{78.71} \\
$\mathrm{ToNER_{xl}}$ & \underline{76.31} & \underline{82.09} & \textbf{79.09} \\
\Xhline{1.2pt}
\end{tabular}
}
\vspace{-0.4cm}
\caption{Results for JNLPBA. \textbf{Bold} and \underline{underline} denote the best and second best scores.}
\label{table:JNLPBA}
\end{table}

\begin{table}[t]
\centering
\resizebox{0.5\textwidth}{!}{
\begin{tabular}{c|ccc|ccc} 
\Xhline{1.2pt} 
 \multirow{2}*{\textbf{Model}} &  \multicolumn{3}{c|}{\textbf{ACE2004}} & \multicolumn{3}{c}{\textbf{ACE2005}} \\
 & {P} & {R} & {F1} & {P} & {R} & {F1} \\ 
 \Xhline{1pt} 
 \citeauthor{yu-etal-2020-named} (\citeyear{yu-etal-2020-named}) & 87.30 & 86.00 & 86.70 & 85.20 & 85.60 & 85.40 \\
 \citeauthor{li-etal-2020-unified} (\citeyear{li-etal-2020-unified}) & 85.05 & 86.32 & 85.98 & \textbf{87.16} & 86.59 & \underline{86.88} \\
 \citeauthor{yan2021unified} (\citeyear{yan2021unified}) & 87.27 & 86.41 & 86.84 & 83.16 & 86.38 & 84.74 \\
 \citeauthor{li2022unified} (\citeyear{li2022unified}) &  87.33 & \underline{87.71} & 87.52 & 85.03 & \textbf{88.62} & 86.79 \\
 \citeauthor{shen-etal-2023-diffusionner} (\citeyear{shen-etal-2023-diffusionner}) & 88.11 & \textbf{88.66} & \textbf{88.39} & 86.15 & 87.72 & \textbf{86.93} \\
 \Xhline{1pt}
$\mathrm{ToNER_{large}}$ & \underline{88.39} & 85.29 & 86.81 & 84.74 & 84.68 & 84.71  \\
$\mathrm{ToNER_{xl}}$ & \textbf{90.03} & 86.24 & \underline{88.09} & \underline{86.66} & \underline{86.71} & 86.68 \\
\Xhline{1.2pt}
\end{tabular}
}
\vspace{-0.4cm}
\caption{Results for ACE2004 and ACE2005. \textbf{Bold} and \underline{underline} denote the best and second best scores.}
\label{table:ACE2004}
\end{table}

\nop{
\begin{table}[t]
\centering
\resizebox{0.45\textwidth}{!}{
\begin{tabular}{c|ccc} 
\Xhline{1.2pt} 
 \multirow{2}*{\textbf{Model}} & \multicolumn{3}{c}{\textbf{CADEC}}  \\
 & {P} & {R} & {F1}\\ 
 \Xhline{1pt} 
 \citeauthor{dai2020effective} (\citeyear{dai2020effective}) & 68.90 & 69.00 & 69.00 \\
 \citeauthor{yan2021unified} (\citeyear{yan2021unified}) & 70.08 & 71.21 & 70.64 \\
 \citeauthor{Fei_Ji_Li_Liu_Ren_Li_2021} (\citeyear{Fei_Ji_Li_Liu_Ren_Li_2021}) & 75.50 & 71.80 & 72.40 \\
 \citeauthor{li-etal-2021-span} (\citeyear{li-etal-2021-span}) & - & - & 69.90 \\
 \citeauthor{li2022unified} (\citeyear{li2022unified}) & 74.09 & 72.35 & \textbf{73.21} \\
 \Xhline{1pt}
 $\mathrm{ToNER_{base}}$ & 64.18 & 69.50 & 66.73  \\
$\mathrm{ToNER}$-$\mathrm{EXP_{base}}$ & 63.15 & 67.17 & 65.10 \\
$\mathrm{ToNER_{large}}$ & 71.05 & 68.35 & 69.68 \\
$\mathrm{ToNER}$-$\mathrm{EXP_{large}}$ & 69.54 & 66.46 & 67.96 \\
\Xhline{1.2pt}
\end{tabular}
}
\vspace{-0.2cm}
\caption{Results for CADEC dataset.}
\label{table:CADEC}
\end{table}
}

\begin{figure}[t]
    \centering
    \includegraphics[width=0.4\textwidth]{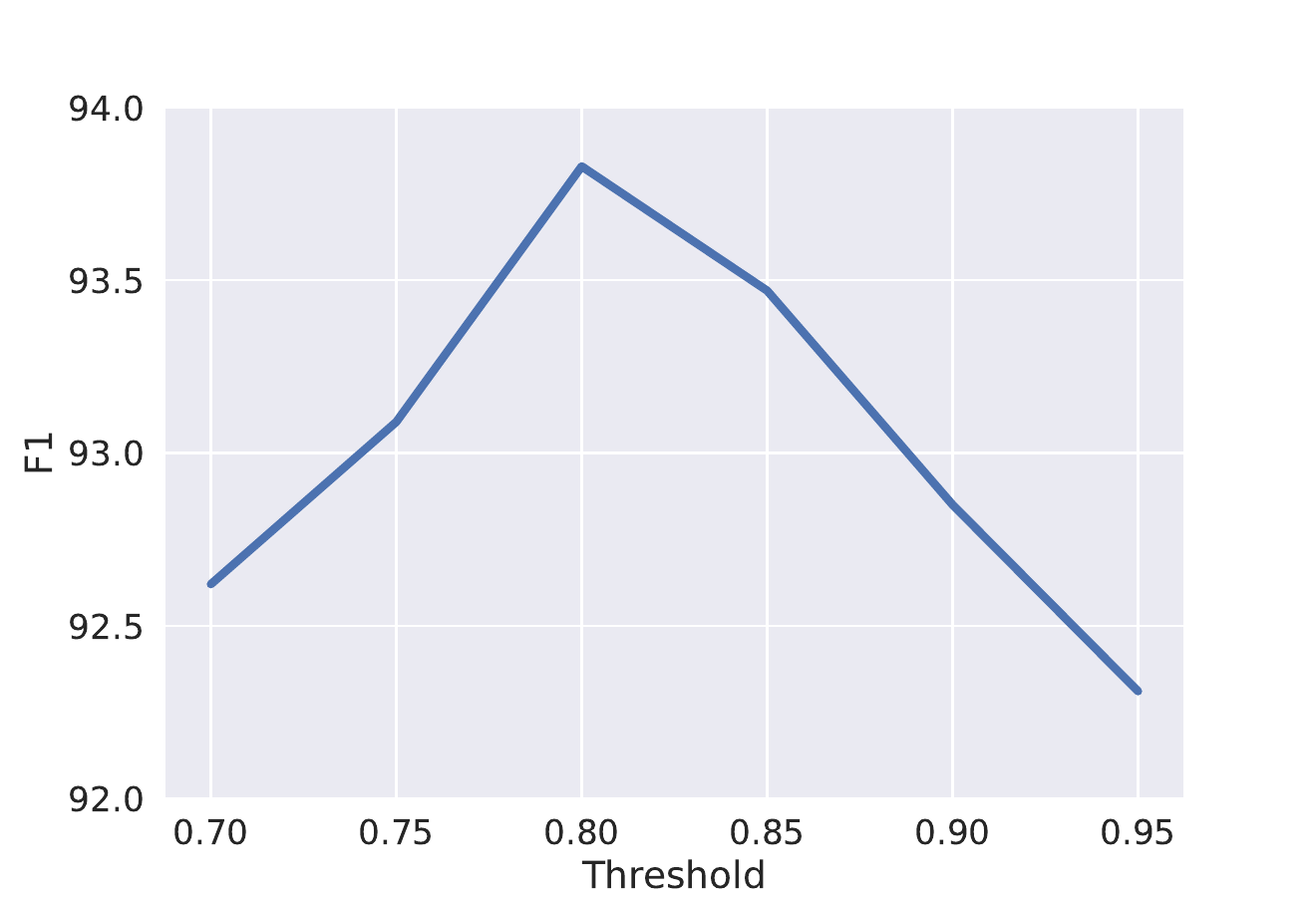}
   \vspace{-0.2cm}
    \caption{The performance of $\mathrm{ToNER}_{\mathrm{large}}$ using different threshold $\delta$ on CoNLL2003.}
    \label{fig:threshold}
\end{figure}

\begin{table}[t]
\centering
\resizebox{0.5\textwidth}{!}{
\begin{tabular}{c|ccc} 
\Xhline{1.2pt} 
 \textbf{Model} & F1 \\ 
 \Xhline{1pt} 
 Flan-T5-large \cite{Flan-T5} & 87.11  \\
Flan-T5-large+$f_\text{TM}$ & 91.18(+4.67\%) \\
Flan-T5-large+$f_\text{TM}$+TC & 92.23(+1.15\%) \\
Flan-T5-large+$f_\text{TM}$+TC+TR ($\mathrm{ToNER_{large}}$) & \textbf{93.83}(+1.73\%) \\
\hline
Flan-T5-xl \cite{Flan-T5} & 89.05  \\
Flan-T5-xl+$f_\text{TM}$ & 91.21(+2.16\%) \\
Flan-T5-xl+$f_\text{TM}$+TC & 92.13(+0.92\%) \\
Flan-T5-xl+$f_\text{TM}$+TC+TR ($\mathrm{ToNER_{xl}}$) & \textbf{93.59}(+1.46\%) \\
\Xhline{1.2pt}
\end{tabular}
}
\vspace{-0.2cm}
\caption{Ablation study results of $\mathrm{{ToNER}_{large}}$ and $\mathrm{{ToNER}_{xl}}$ on CoNLL2003. $f_\text{Tm}$ means the entity type matching model in Section \ref{sec:tm}. TC means the entity type classification for fine-tuning Encoder in Section \ref{sec:tc}. TR means the auxiliary entity type recognition task in Section \ref{sec:tr}.}
\label{table:ablation}
\end{table}

\subsection{Ablation Study}
In order to further justify the effectiveness of each component we propose in ToNER, we compared ToNER with its ablated variants. Specifically, the basic variant only uses the generative model Flan-T5-large and Flan-T5-xl. Then, we added the type matching model $f_\text{TM}$, the type classification (TC) task and the auxiliary of type recognition (TR) into this basic variant in turn. Due to space limitation, table \ref{table:ablation} only lists the performance of $\mathrm{{ToNER}_{large}}$ and $\mathrm{{ToNER}_{xl}}$ along with their corresponding three ablated variants on CoNLL2003. As well, each variant's performance improvement rate relative to the preceding variant is also listed. This table's results obviously show that either $f_\text{TM}$, TC or TR can improve ToNER's performance. The ablation studies on other datasets also support this conclusion.

\subsection{Threshold Selection of Entity Type Matching}
\begin{figure*}[t]
    \centering
    \subfigure[CoNLL2003]{\includegraphics[width=0.19\textwidth]{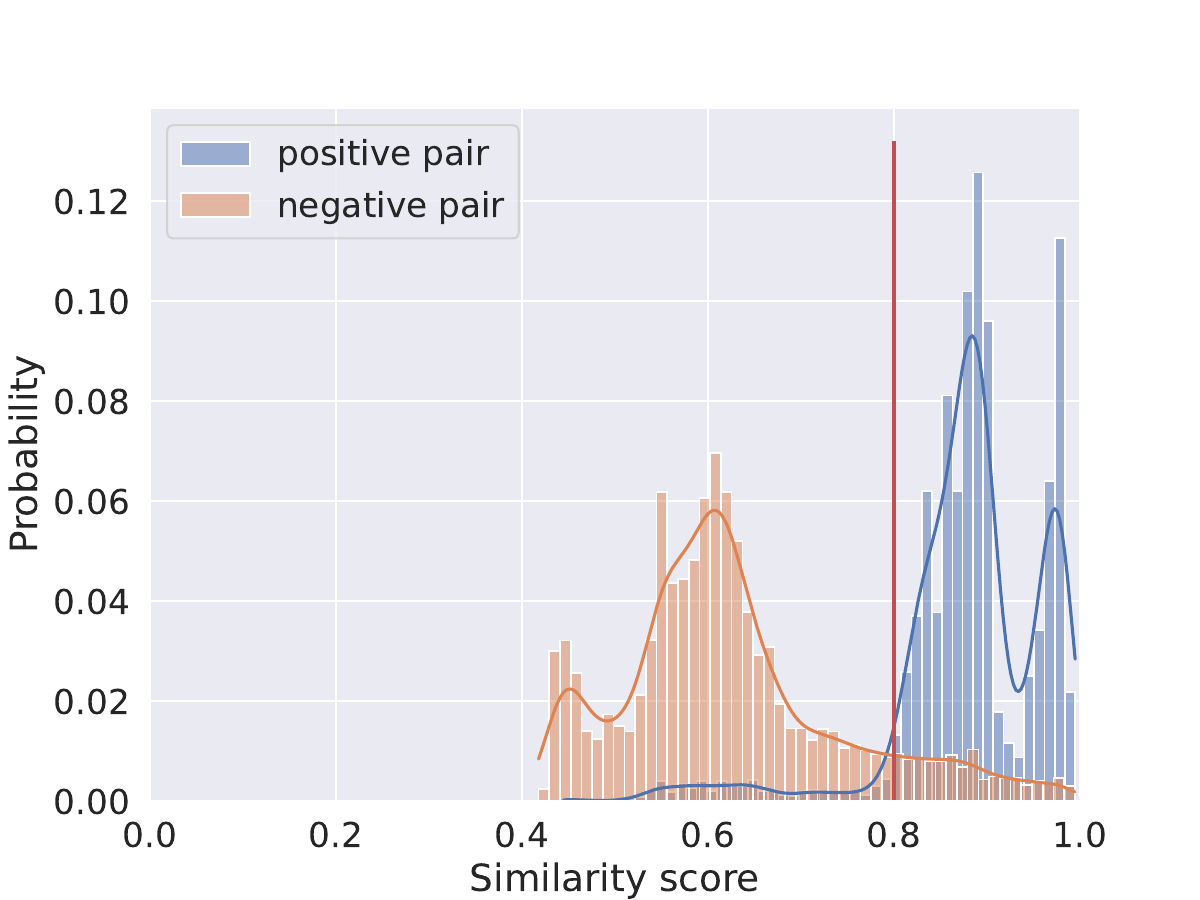}}
    \subfigure[OntoNotes 5.0]{\includegraphics[width=0.19\textwidth]{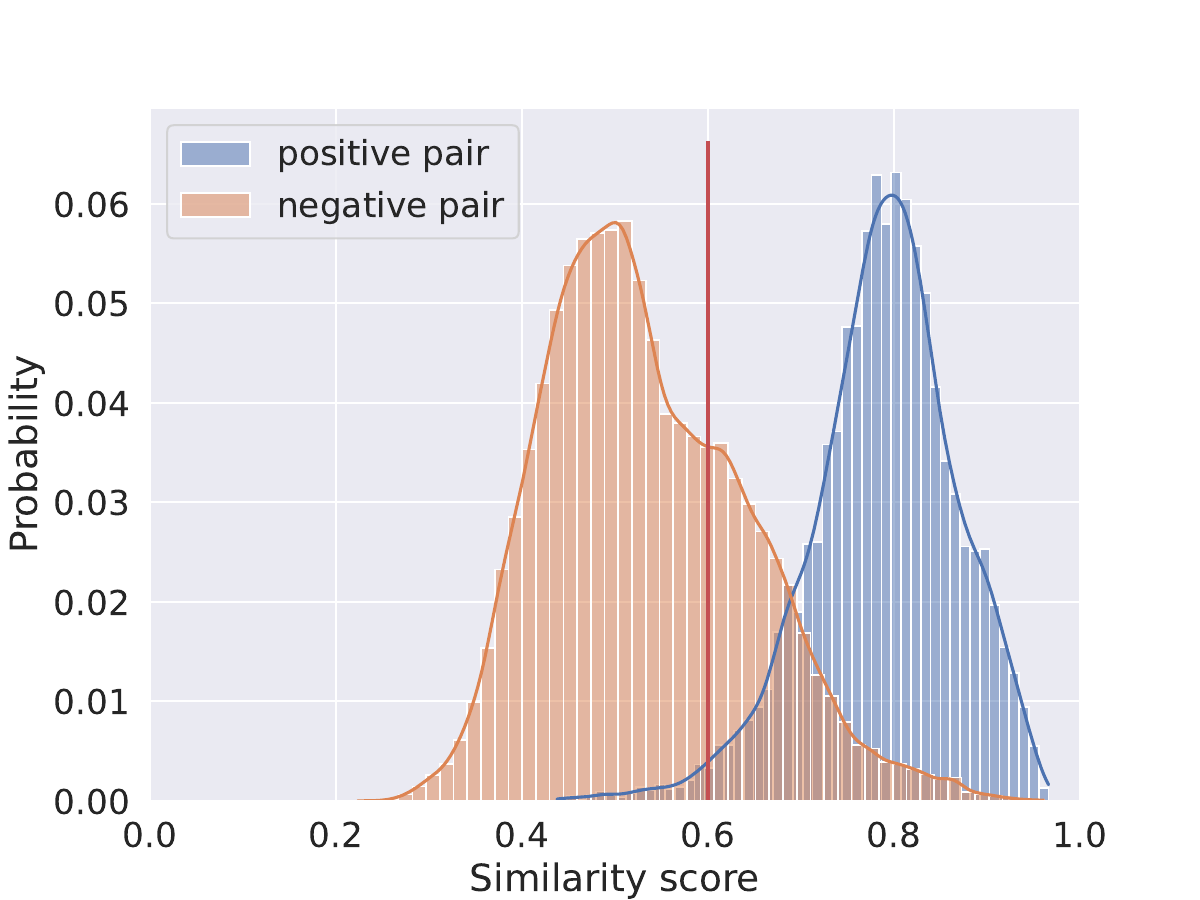}}
    \subfigure[JNLPBA]{\includegraphics[width=0.19\textwidth]{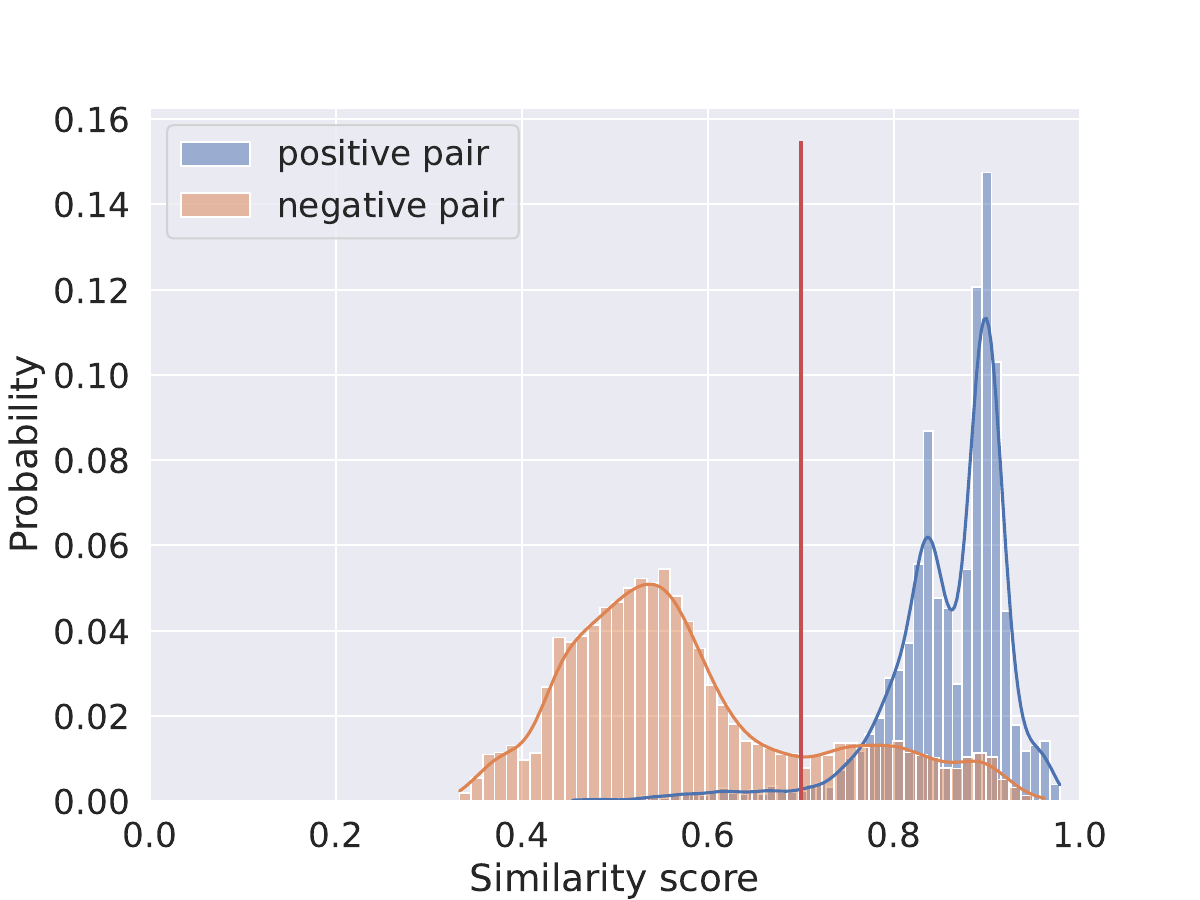}}
    \subfigure[ACE2004]{\includegraphics[width=0.19\textwidth]{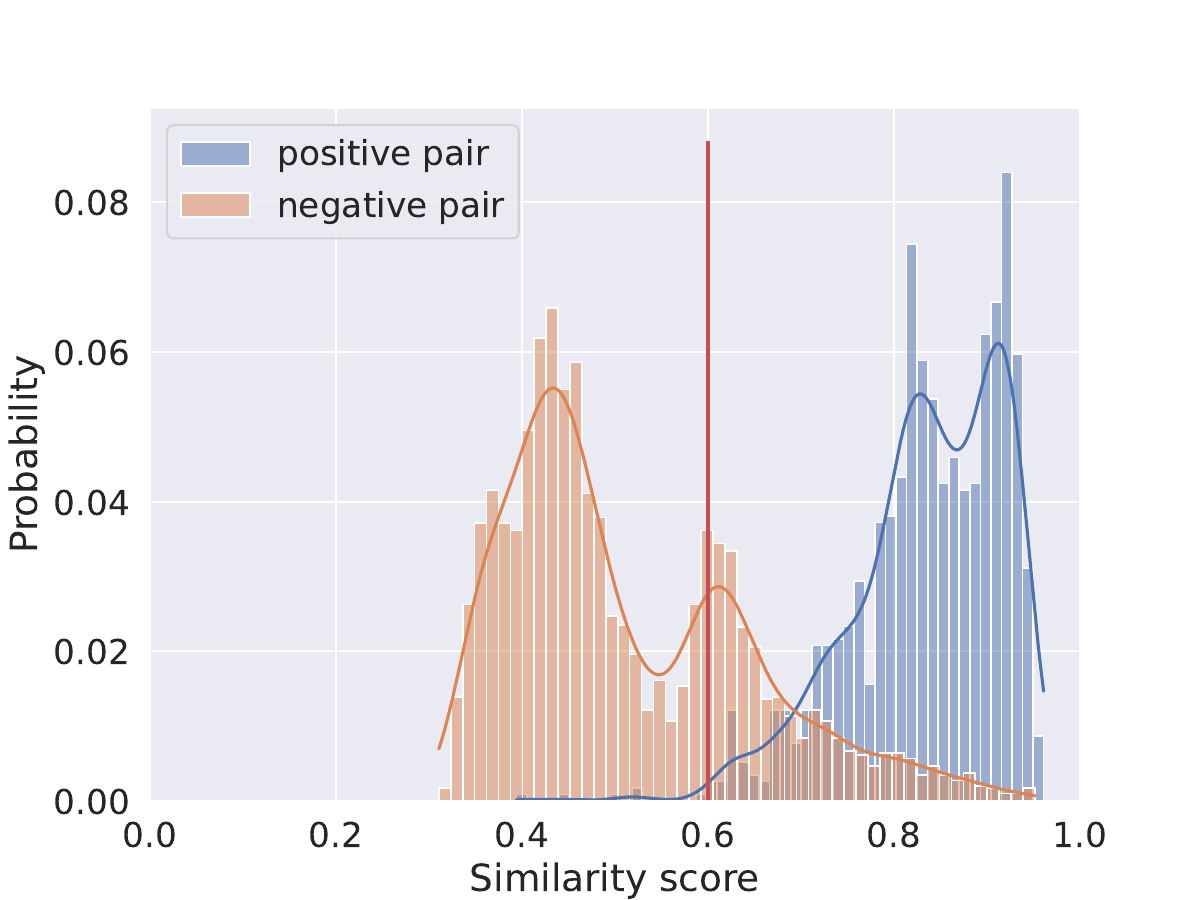}}
    \subfigure[ACE2005]{\includegraphics[width=0.19\textwidth]{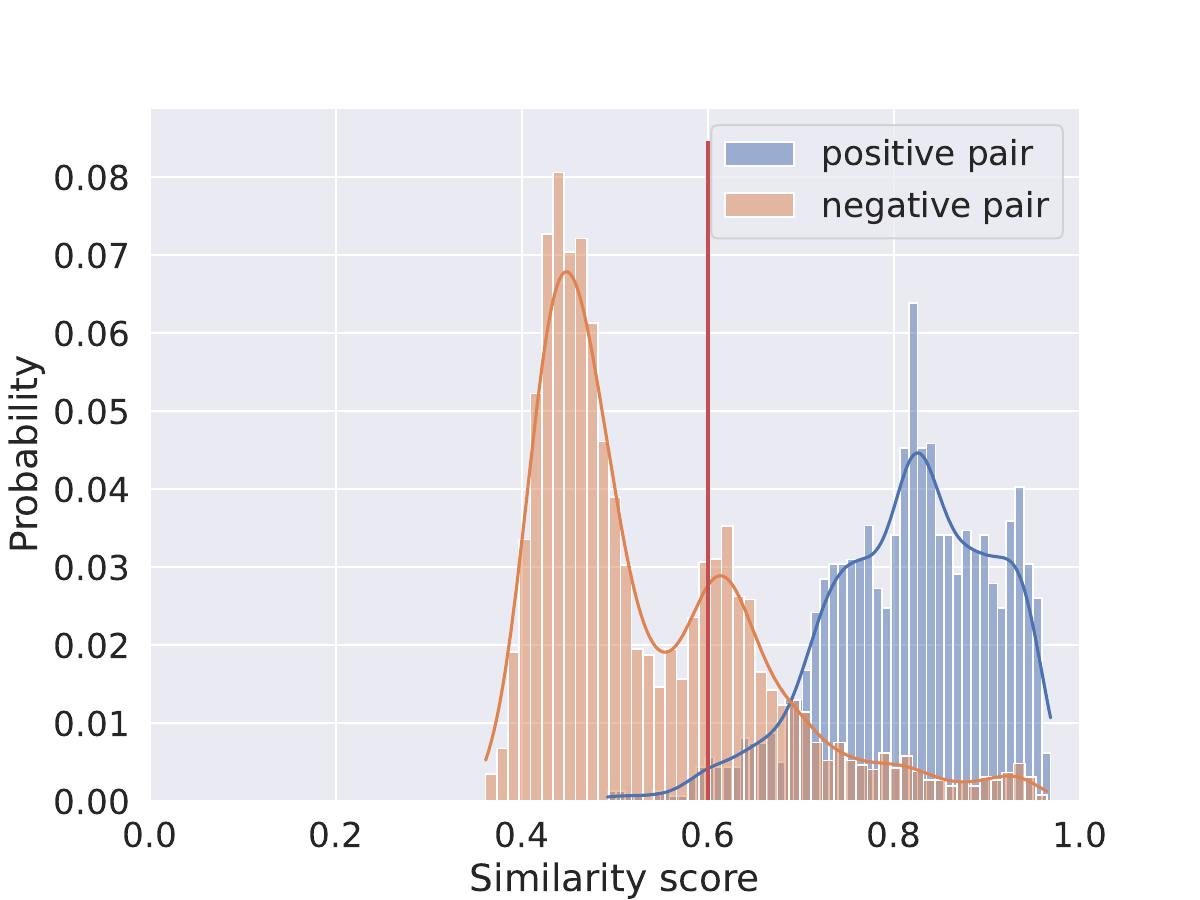}}
    \vspace{-0.2cm}
    \caption{Similarity score distributions of all text-type pairs computed by the fine-tuned type matching model. The \textcolor{red}{red} vertical line represents the threshold $\delta$ we selected. 
    }
    \label{fig:matching_model}
\end{figure*}
We also investigated the type matching model $f_\text{TM}$'s capability of discriminating between the positive text-type pairs and the negative text-type pairs, since it is a key part of ToNER to improve NER performance. Figure \ref{fig:matching_model} displays the distributions of all text-type pairs' similarity score's in the five datasets, which were computed by the fine-tuned $f_\text{TM}$ according to Eq. \ref{eq:score}. The distributions show that $f_\text{TM}$ can well discriminate between the positive pairs and the negative pairs, based on which we can select the best threshold $\delta$, as shown in the five sub-figures. 

We also tested $\delta$'s impact on ToNER's performance. Figure \ref{fig:threshold} depicts $\mathrm{ToNER}_{\mathrm{large}}$'s F1 score on CoNLL2003 as $\delta$ varies from 0.7 to 0.95. It shows that $\delta=0.8$ is the best setting for this dataset.

\subsection{Impacts of CoT-style Explanations and Model Size}
\begin{figure*}[ht]
    \centering
    \subfigure[CoNLL2003]{\includegraphics[width=0.3\textwidth]{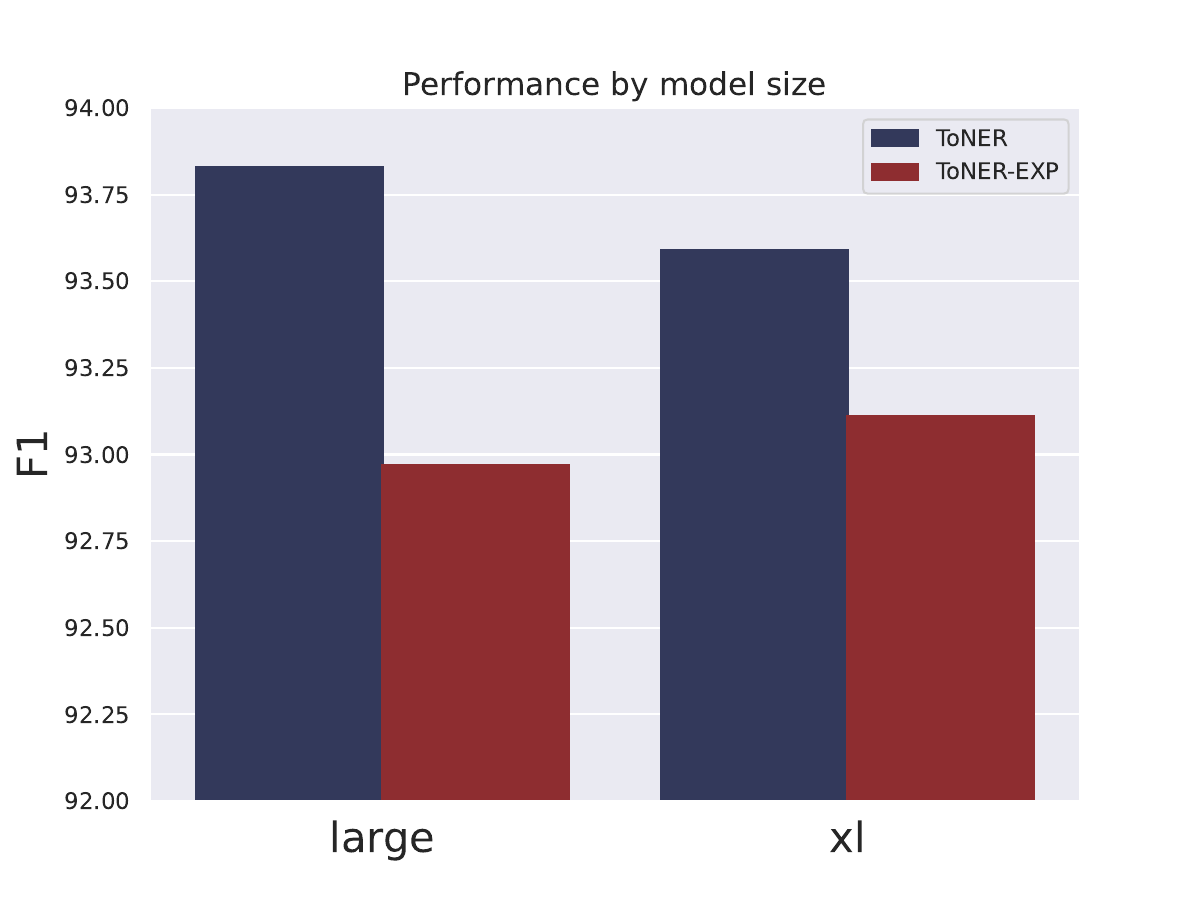}}
    \subfigure[OntoNotes 5.0]{\includegraphics[width=0.3\textwidth]{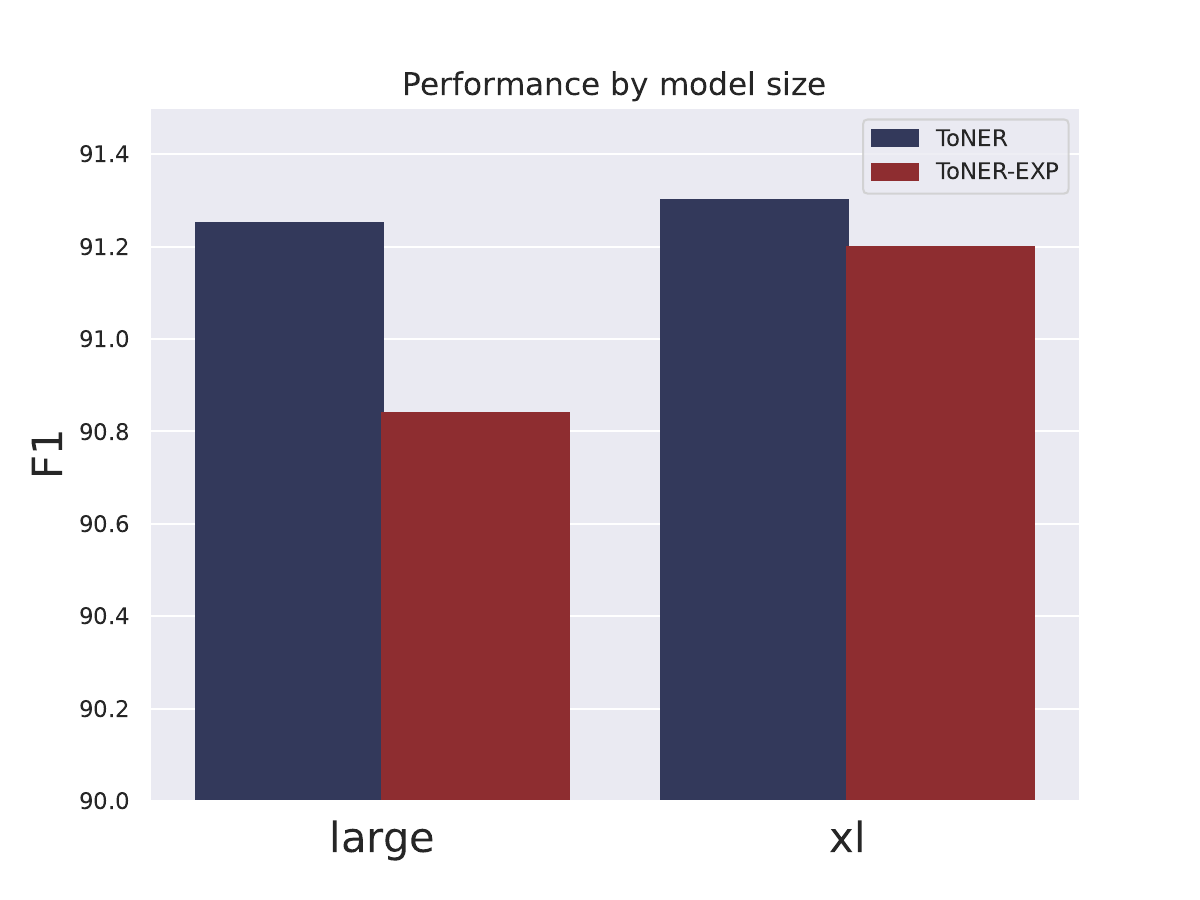}}
    \subfigure[JNLPBA]{\includegraphics[width=0.3\textwidth]{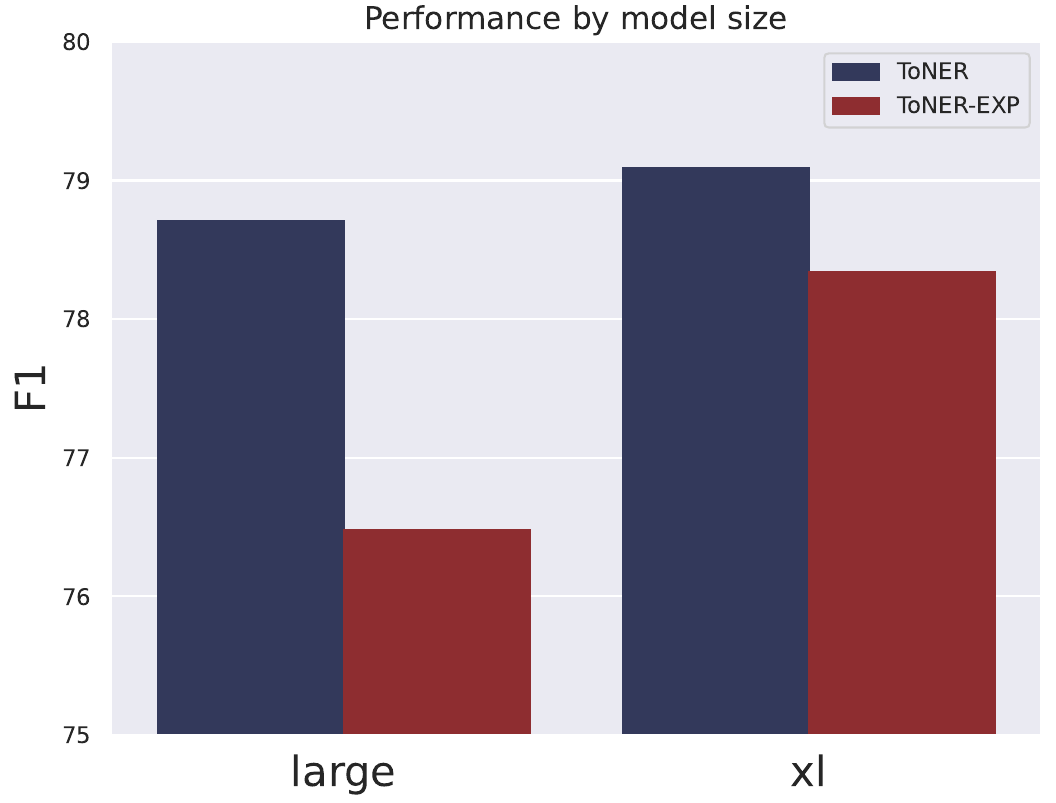}}
    \subfigure[ACE2004]{\includegraphics[width=0.3\textwidth]{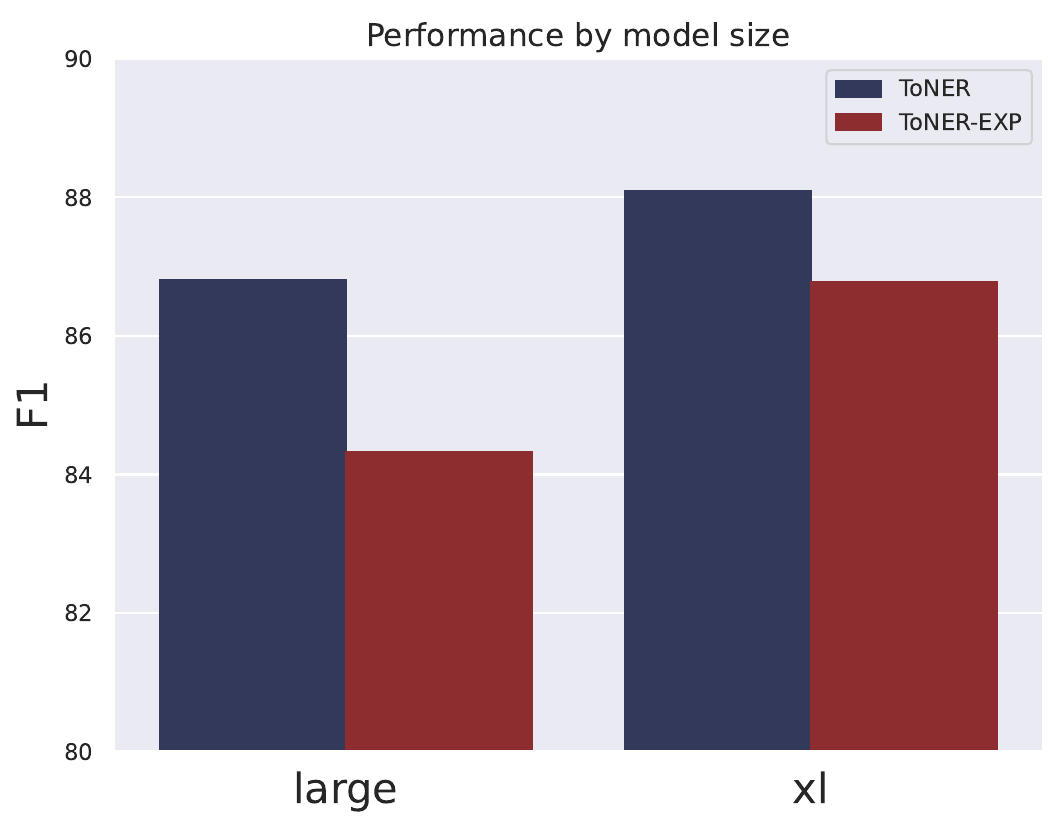}}
    \subfigure[ACE2005]{\includegraphics[width=0.3\textwidth]{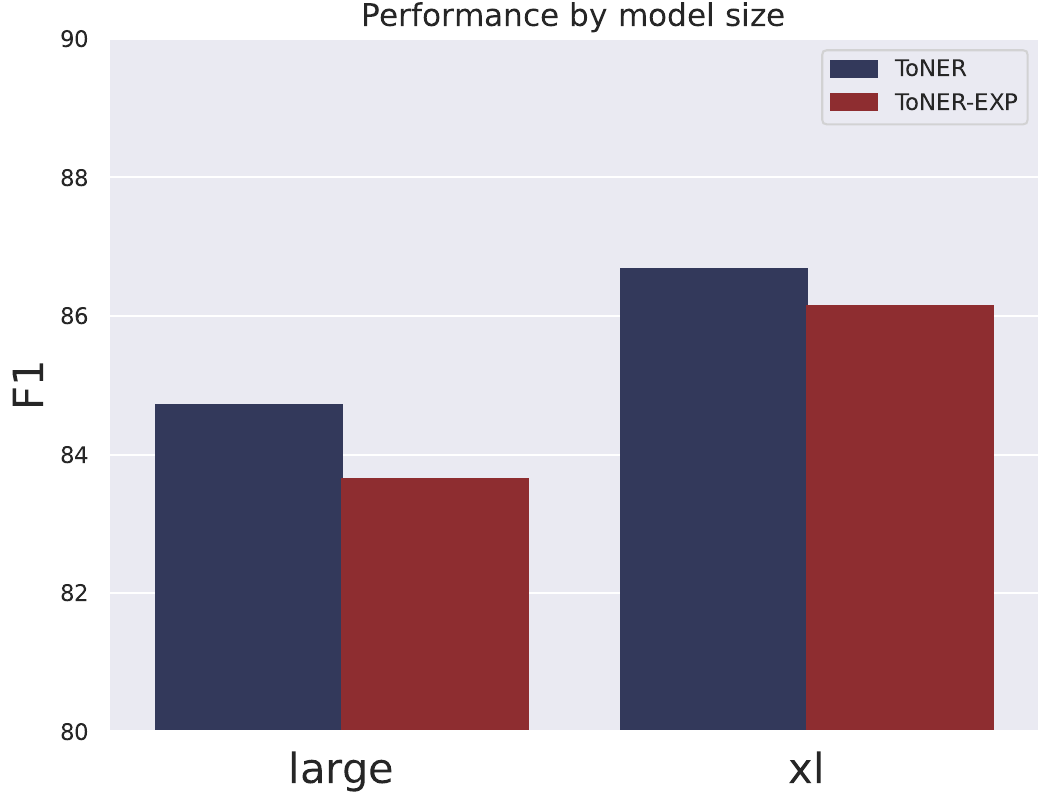}}
    \vspace{-0.2cm}
    \caption{The performance of ToNER and ToNER-EXP with different model size on different datasets.}
    \label{fig:exp}
\end{figure*}

\begin{table}[t]
\centering
\resizebox{0.45\textwidth}{!}{
\begin{tabular}{c|lll} 
\Xhline{1.2pt} 
 \multirow{2}*{\textbf{Model}} &  \multicolumn{3}{c}{\textbf{CoNLL2003}}  \\
 & {P} & {R} & {F1}\\ 
 \Xhline{1pt} 
$\mathrm{ToNER_{large}}$ & \underline{93.55} & \textbf{94.11} & \textbf{93.83} \\
$\mathrm{ToNER_{large}}$-$\mathrm{EXP}$ & 93.18 (-0.40\%) & 92.77 (-1.42\%) & 92.97 (-0.92\%) \\
\hline
$\mathrm{ToNER_{xl}}$ & 93.53 & \underline{93.65} & 93.59 \\
$\mathrm{ToNER_{xl}}$-$\mathrm{EXP}$ & 93.10 (-0.46\%) & 93.13 (-0.56\%) & 93.11 (-0.51\%) \\
\hline
$\mathrm{ToNER_{xxl}}$ & 92.74 & 92.28 & 92.52 \\
$\mathrm{ToNER_{xxl}}$-$\mathrm{EXP}$ & \textbf{93.93} (+1.28\%) & 93.47 (+1.29\%) & \underline{93.70} (+1.28\%)\\
\Xhline{1.2pt}
\end{tabular}
}
\vspace{-0.2cm}
\caption{ToNER and ToNER-EXP's performance on CoNLL2003. \textbf{Bold} and \underline{underline} denote the best and second best scores.
}
\label{table:model_size}
\end{table}

It has been found that CoT only has a positive effect on sufficiently large models (typically containing 10B or more parameters) but not on small models \cite{COT}, since CoT is an emergent ability \cite{Emergent}.
In order to explore the impact of adding CoT explanations to achieve NER task, we compared the performance of ToNER and ToNER-EXP in the \texttt{large} and \texttt{xl} setting, as shown in Figure \ref{fig:exp}. For these datasets, the model performance improvement of ToNER-EXP is higher than that of ToNER when the model size changes from \texttt{large} to \texttt{xl}. This suggests that the model's ability to generate CoT-style explanations may gradually increase as the number of parameters increases. Increasing model size not only helps direct NER performance but also improves CoT-style explanation and helps generate NER results indirectly.

In order to further explore the effect of further increasing the model size, we selected the CoNLL2003 dataset for further exploration. Specifically, besides $\mathrm{ToNER}_{\mathrm{large}}$ and $\mathrm{ToNER}_{\mathrm{xl}}$, we further considered $\mathrm{ToNER}_{\mathrm{xxl}}$, which use the Flan-T5 versions of 780M, 3B, 11B parameters, respectively. Table \ref{table:model_size} lists their performance on CoNLL2003, including the relative performance improvement rate of each version's ToNER-EXP to its corresponding ToNER. From the table we find that, only for the $\mathrm{xxl}$ version, ToNER-EXP can improve NER performance, verifying the previous finding that CoT's effectiveness on large models rather than small models. When the generative model's scale is large enough, the CoT-style explanations can fine-tune the model to better utilize its rich knowledge to understand the input texts correctly, thus improving NER performance further.

Another interesting observation is that, $\mathrm{ToNER}_{\mathrm{large}}$ has the best R score and F1 score although it only has the fewest parameters. Instead, further increasing the parameters of the generative model degrades the performance. This could be attributed to that, fine-tuning a large model for optimal performance necessitates a broader and larger dataset. The available dataset can not meet this requirement. Our experiments on other datasets have the similar results.

\section{Conclusion}
In this paper, we propose a novel NER framework \emph{ToNER} based on a generative language model. In ToNER we further employ an entity type matching model to discover the entity types mostly likely to appear in the sentence, which are input into the generative model for more concentrations. Additional classification learning objectives are also designed to fine-tune the generative model, to improve ToNER's performance further. At the same time, we also explored the impact of generating CoT-style explanations for model outputs. Our experiments on five NER datasets illustrate the advantages of ToNER over the previous models.

\section*{Limitations}
We only explore the feasibility of the Encoder-Decoder architecture model as a base model for ToNER. For generative language models, more existing options are based on Decoder-only. This limitation highlights the potential for future work to explore different model architectures to understand named entity recognition.

\section*{Acknowledgements}
This work was supported by the Chinese NSF Major Research Plan (No.92270121), Youth Fund (No.62102095), Shanghai Science and Technology Innovation Action Plan (No.21511100401). The computations in this research were performed using the CFFF platform of Fudan University.

\section*{Bibliographical
 References}\label{sec:reference}

\bibliographystyle{lrec-coling2024-natbib}
\bibliography{lrec-coling2024-example}

\begin{thebibliography}{5}
\expandafter\ifx\csname natexlab\endcsname\relax\def\natexlab#1{#1}\fi

\bibitem[{Alexis et~al.(2004)Alexis, Stephanie, Shudong, and
  Ramez}]{ace2004_lr}
Alexis, Mitchell and Stephanie, Strassel and Shudong, Huang and Ramez, Zakhary.
  2004.
\newblock \emph{ACE 2004 Multilingual Training Corpus}.
\newblock European Language Resources Association (ELRA), ISLRN
  \href{https://www.islrn.org/resources/789-870-824-708-5}{789-870-824-708-5}.

\bibitem[{Collier et~al.(2004)Collier, Ohta, Tsuruoka, Tateisi, and
  Kim}]{JNLPBA_lr}
Collier, Nigel and Ohta, Tomoko and Tsuruoka, Yoshimasa and Tateisi, Yuka and
  Kim, Jin-Dong. 2004.
\newblock \emph{Introduction to the Bio-entity Recognition Task at {JNLPBA}}.
\newblock COLING.
\newblock PID
  \href{https://aclanthology.org/W04-1213}{https://aclanthology.org/W04-1213}.

\bibitem[{Hovy et~al.(2006)Hovy, Marcus, Palmer, Ramshaw, and
  Weischedel}]{ontonotes_lr}
Hovy, Eduard and Marcus, Mitchell and Palmer, Martha and Ramshaw, Lance and
  Weischedel, Ralph. 2006.
\newblock \emph{{O}nto{N}otes: The 90{\%} Solution}.
\newblock Association for Computational Linguistics, ISLRN
  \href{https://www.islrn.org/resources/151-738-649-048-2}{151-738-649-048-2}.

\bibitem[{Tjong Kim~Sang and De~Meulder(2003)}]{conll2003_lr}
Tjong Kim Sang, Erik F. and De Meulder, Fien. 2003.
\newblock \emph{Introduction to the {C}o{NLL}-2003 Shared Task:
  Language-Independent Named Entity Recognition}.
\newblock PID
  \href{https://aclanthology.org/W03-0419}{https://aclanthology.org/W03-0419}.

\bibitem[{Walker et~al.(2006)Walker, Strassel, Medero, and Maeda}]{ace2005_lr}
Walker, Christopher and Strassel, Stephanie and Medero, Julie and Maeda,
  Kazuaki. 2006.
\newblock \emph{ACE 2005 Multilingual Training Corpus}.
\newblock European Language Resources Association (ELRA), ISLRN
  \href{https://www.islrn.org/resources/458-031-085-383-4}{458-031-085-383-4}.

\end{thebibliography}


\begin{thebibliography}{48}
\expandafter\ifx\csname natexlab\endcsname\relax\def\natexlab#1{#1}\fi

\bibitem[{Banerjee et~al.(2019)Banerjee, Chakraborty, Tripathi, Gupta, and
  Kumar}]{banerjee2019information}
Partha~Sarathy Banerjee, Baisakhi Chakraborty, Deepak Tripathi, Hardik Gupta,
  and Sourabh~S. Kumar. 2019.
\newblock \href {https://doi.org/10.1007/s11277-019-06501-z} {A information
  retrieval based on question and answering and ner for unstructured
  information without using sql}.
\newblock \emph{Wirel. Pers. Commun.}, 108(3):1909–1931.

\bibitem[{Brown et~al.(2020)Brown, Mann, Ryder, Subbiah, Kaplan, Dhariwal,
  Neelakantan, Shyam, Sastry, Askell, Agarwal, Herbert-Voss, Krueger, Henighan,
  Child, Ramesh, Ziegler, Wu, Winter, Hesse, Chen, Sigler, Litwin, Gray, Chess,
  Clark, Berner, McCandlish, Radford, Sutskever, and Amodei}]{GPT-3}
Tom Brown, Benjamin Mann, Nick Ryder, Melanie Subbiah, Jared~D Kaplan, Prafulla
  Dhariwal, Arvind Neelakantan, Pranav Shyam, Girish Sastry, Amanda Askell,
  Sandhini Agarwal, Ariel Herbert-Voss, Gretchen Krueger, Tom Henighan, Rewon
  Child, Aditya Ramesh, Daniel Ziegler, Jeffrey Wu, Clemens Winter, Chris
  Hesse, Mark Chen, Eric Sigler, Mateusz Litwin, Scott Gray, Benjamin Chess,
  Jack Clark, Christopher Berner, Sam McCandlish, Alec Radford, Ilya Sutskever,
  and Dario Amodei. 2020.
\newblock \href
  {https://proceedings.neurips.cc/paper_files/paper/2020/file/1457c0d6bfcb4967418bfb8ac142f64a-Paper.pdf}
  {Language models are few-shot learners}.
\newblock In \emph{Advances in Neural Information Processing Systems},
  volume~33, pages 1877--1901. Curran Associates, Inc.

\bibitem[{Chen et~al.(2021)Chen, Li, Deng, Tan, Xu, Huang, Si, Chen, and
  Zhang}]{chen2021lightner}
Xiang Chen, Lei Li, Shumin Deng, Chuanqi Tan, Changliang Xu, Fei Huang, Luo Si,
  Huajun Chen, and Ningyu Zhang. 2021.
\newblock \href {http://arxiv.org/abs/2109.00720} {Lightner: {A} lightweight
  tuning paradigm for low-resource {NER} via pluggable prompting}.
\newblock \emph{CoRR}, abs/2109.00720.

\bibitem[{{Chung} et~al.(2022){Chung}, {Hou}, {Longpre}, {Zoph}, {Tay},
  {Fedus}, {Li}, {Wang}, {Dehghani}, {Brahma}, {Webson}, {Gu}, {Dai}, {Suzgun},
  {Chen}, {Chowdhery}, {Castro-Ros}, {Pellat}, {Robinson}, {Valter}, {Narang},
  {Mishra}, {Yu}, {Zhao}, {Huang}, {Dai}, {Yu}, {Petrov}, {Chi}, {Dean},
  {Devlin}, {Roberts}, {Zhou}, {Le}, and {Wei}}]{Flan-T5}
Hyung~Won {Chung}, Le~{Hou}, Shayne {Longpre}, Barret {Zoph}, Yi~{Tay}, William
  {Fedus}, Yunxuan {Li}, Xuezhi {Wang}, Mostafa {Dehghani}, Siddhartha
  {Brahma}, Albert {Webson}, Shixiang~Shane {Gu}, Zhuyun {Dai}, Mirac {Suzgun},
  Xinyun {Chen}, Aakanksha {Chowdhery}, Alex {Castro-Ros}, Marie {Pellat},
  Kevin {Robinson}, Dasha {Valter}, Sharan {Narang}, Gaurav {Mishra}, Adams
  {Yu}, Vincent {Zhao}, Yanping {Huang}, Andrew {Dai}, Hongkun {Yu}, Slav
  {Petrov}, Ed~H. {Chi}, Jeff {Dean}, Jacob {Devlin}, Adam {Roberts}, Denny
  {Zhou}, Quoc~V. {Le}, and Jason {Wei}. 2022.
\newblock \href {https://doi.org/10.48550/arXiv.2210.11416} {{Scaling
  Instruction-Finetuned Language Models}}.
\newblock \emph{arXiv e-prints}, page arXiv:2210.11416.

\bibitem[{Cui et~al.(2021)Cui, Wu, Liu, Yang, and Zhang}]{cui2021template}
Leyang Cui, Yu~Wu, Jian Liu, Sen Yang, and Yue Zhang. 2021.
\newblock \href {https://doi.org/10.18653/v1/2021.findings-acl.161}
  {Template-based named entity recognition using {BART}}.
\newblock In \emph{Findings of the Association for Computational Linguistics:
  ACL-IJCNLP 2021}, pages 1835--1845, Online. Association for Computational
  Linguistics.

\bibitem[{Dai et~al.(2020)Dai, Karimi, Hachey, and Paris}]{dai2020effective}
Xiang Dai, Sarvnaz Karimi, Ben Hachey, and Cecile Paris. 2020.
\newblock \href {https://doi.org/10.18653/v1/2020.acl-main.520} {An effective
  transition-based model for discontinuous {NER}}.
\newblock In \emph{Proceedings of the 58th Annual Meeting of the Association
  for Computational Linguistics}, pages 5860--5870, Online. Association for
  Computational Linguistics.

\bibitem[{Devlin et~al.(2019)Devlin, Chang, Lee, and Toutanova}]{BERT}
Jacob Devlin, Ming-Wei Chang, Kenton Lee, and Kristina Toutanova. 2019.
\newblock \href {https://doi.org/10.18653/v1/N19-1423} {{BERT}: Pre-training of
  deep bidirectional transformers for language understanding}.
\newblock In \emph{Proceedings of the 2019 Conference of the North {A}merican
  Chapter of the Association for Computational Linguistics: Human Language
  Technologies, Volume 1 (Long and Short Papers)}, pages 4171--4186,
  Minneapolis, Minnesota. Association for Computational Linguistics.

\bibitem[{Fei et~al.(2021)Fei, Ji, Li, Liu, Ren, and
  Li}]{Fei_Ji_Li_Liu_Ren_Li_2021}
Hao Fei, Donghong Ji, Bobo Li, Yijiang Liu, Yafeng Ren, and Fei Li. 2021.
\newblock \href {https://doi.org/10.1609/aaai.v35i14.17513} {Rethinking
  boundaries: End-to-end recognition of discontinuous mentions with pointer
  networks}.
\newblock \emph{Proceedings of the AAAI Conference on Artificial Intelligence},
  35(14):12785--12793.

\bibitem[{Fu et~al.(2021)Fu, Huang, and Liu}]{fu-etal-2021-spanner}
Jinlan Fu, Xuanjing Huang, and Pengfei Liu. 2021.
\newblock \href {https://doi.org/10.18653/v1/2021.acl-long.558} {{S}pan{NER}:
  Named entity re-/recognition as span prediction}.
\newblock In \emph{Proceedings of the 59th Annual Meeting of the Association
  for Computational Linguistics and the 11th International Joint Conference on
  Natural Language Processing (Volume 1: Long Papers)}, pages 7183--7195,
  Online. Association for Computational Linguistics.

\bibitem[{Gao et~al.(2021)Gao, Yao, and Chen}]{SimCSE}
Tianyu Gao, Xingcheng Yao, and Danqi Chen. 2021.
\newblock \href {https://doi.org/10.18653/v1/2021.emnlp-main.552} {{S}im{CSE}:
  Simple contrastive learning of sentence embeddings}.
\newblock In \emph{Proceedings of the 2021 Conference on Empirical Methods in
  Natural Language Processing}, pages 6894--6910, Online and Punta Cana,
  Dominican Republic. Association for Computational Linguistics.

\bibitem[{Katiyar and Cardie(2018)}]{katiyar-cardie-2018-nested}
Arzoo Katiyar and Claire Cardie. 2018.
\newblock \href {https://doi.org/10.18653/v1/N18-1079} {Nested named entity
  recognition revisited}.
\newblock In \emph{Proceedings of the 2018 Conference of the North {A}merican
  Chapter of the Association for Computational Linguistics: Human Language
  Technologies, Volume 1 (Long Papers)}, pages 861--871, New Orleans,
  Louisiana. Association for Computational Linguistics.

\bibitem[{Lee et~al.(2019)Lee, Yoon, Kim, Kim, Kim, So, and
  Kang}]{10.1093/bioinformatics/btz682}
Jinhyuk Lee, Wonjin Yoon, Sungdong Kim, Donghyeon Kim, Sunkyu Kim, Chan~Ho So,
  and Jaewoo Kang. 2019.
\newblock \href {https://doi.org/10.1093/bioinformatics/btz682} {{BioBERT: a
  pre-trained biomedical language representation model for biomedical text
  mining}}.
\newblock \emph{Bioinformatics}.

\bibitem[{Lewis et~al.(2020)Lewis, Liu, Goyal, Ghazvininejad, Mohamed, Levy,
  Stoyanov, and Zettlemoyer}]{BART}
Mike Lewis, Yinhan Liu, Naman Goyal, Marjan Ghazvininejad, Abdelrahman Mohamed,
  Omer Levy, Veselin Stoyanov, and Luke Zettlemoyer. 2020.
\newblock \href {https://doi.org/10.18653/v1/2020.acl-main.703} {{BART}:
  Denoising sequence-to-sequence pre-training for natural language generation,
  translation, and comprehension}.
\newblock In \emph{Proceedings of the 58th Annual Meeting of the Association
  for Computational Linguistics}, pages 7871--7880, Online. Association for
  Computational Linguistics.

\bibitem[{Li et~al.(2021)Li, Lin, Zhang, and Ji}]{li-etal-2021-span}
Fei Li, ZhiChao Lin, Meishan Zhang, and Donghong Ji. 2021.
\newblock \href {https://doi.org/10.18653/v1/2021.acl-long.372} {A span-based
  model for joint overlapped and discontinuous named entity recognition}.
\newblock In \emph{Proceedings of the 59th Annual Meeting of the Association
  for Computational Linguistics and the 11th International Joint Conference on
  Natural Language Processing (Volume 1: Long Papers)}, pages 4814--4828,
  Online. Association for Computational Linguistics.

\bibitem[{Li et~al.(2022{\natexlab{a}})Li, Sun, Han, and Li}]{NER}
Jing Li, Aixin Sun, Jianglei Han, and Chenliang Li. 2022{\natexlab{a}}.
\newblock \href {https://doi.org/10.1109/TKDE.2020.2981314} {A survey on deep
  learning for named entity recognition}.
\newblock \emph{IEEE Transactions on Knowledge and Data Engineering},
  34(1):50--70.

\bibitem[{Li et~al.(2022{\natexlab{b}})Li, Fei, Liu, Wu, Zhang, Teng, Ji, and
  Li}]{li2022unified}
Jingye Li, Hao Fei, Jiang Liu, Shengqiong Wu, Meishan Zhang, Chong Teng,
  Donghong Ji, and Fei Li. 2022{\natexlab{b}}.
\newblock \href {https://doi.org/https://doi.org/10.1609/aaai.v36i10.21344}
  {Unified named entity recognition as word-word relation classification}.
\newblock In \emph{Proceedings of the AAAI Conference on Artificial
  Intelligence}, pages 10965--10973.

\bibitem[{Li et~al.(2020)Li, Feng, Meng, Han, Wu, and
  Li}]{li-etal-2020-unified}
Xiaoya Li, Jingrong Feng, Yuxian Meng, Qinghong Han, Fei Wu, and Jiwei Li.
  2020.
\newblock \href {https://doi.org/10.18653/v1/2020.acl-main.519} {A unified
  {MRC} framework for named entity recognition}.
\newblock In \emph{Proceedings of the 58th Annual Meeting of the Association
  for Computational Linguistics}, pages 5849--5859, Online. Association for
  Computational Linguistics.

\bibitem[{Li and Qian(2023)}]{li2023type}
Yongqi Li and Tieyun Qian. 2023.
\newblock Type-aware decomposed framework for few-shot named entity
  recognition.
\newblock \emph{arXiv preprint arXiv:2302.06397}.

\bibitem[{{Li} et~al.(2023){Li}, {Zhang}, {Zhang}, {Long}, {Xie}, and
  {Zhang}}]{gte}
Zehan {Li}, Xin {Zhang}, Yanzhao {Zhang}, Dingkun {Long}, Pengjun {Xie}, and
  Meishan {Zhang}. 2023.
\newblock \href {https://doi.org/10.48550/arXiv.2308.03281} {{Towards General
  Text Embeddings with Multi-stage Contrastive Learning}}.
\newblock \emph{arXiv e-prints}, page arXiv:2308.03281.

\bibitem[{Lin et~al.(2019)Lin, Lu, Han, and Sun}]{lin-etal-2019-sequence}
Hongyu Lin, Yaojie Lu, Xianpei Han, and Le~Sun. 2019.
\newblock \href {https://doi.org/10.18653/v1/P19-1511} {Sequence-to-nuggets:
  Nested entity mention detection via anchor-region networks}.
\newblock In \emph{Proceedings of the 57th Annual Meeting of the Association
  for Computational Linguistics}, pages 5182--5192, Florence, Italy.
  Association for Computational Linguistics.

\bibitem[{Loshchilov and Hutter(2017)}]{adamw}
Ilya Loshchilov and Frank Hutter. 2017.
\newblock \href {http://arxiv.org/abs/1711.05101} {Fixing weight decay
  regularization in adam}.
\newblock \emph{CoRR}, abs/1711.05101.

\bibitem[{Lu and Roth(2015)}]{lu2015joint}
Wei Lu and Dan Roth. 2015.
\newblock \href {https://doi.org/10.18653/v1/D15-1102} {Joint mention
  extraction and classification with mention hypergraphs}.
\newblock In \emph{Proceedings of the 2015 Conference on Empirical Methods in
  Natural Language Processing}, pages 857--867, Lisbon, Portugal. Association
  for Computational Linguistics.

\bibitem[{Lu et~al.(2022)Lu, Liu, Dai, Xiao, Lin, Han, Sun, and
  Wu}]{lu2022unified}
Yaojie Lu, Qing Liu, Dai Dai, Xinyan Xiao, Hongyu Lin, Xianpei Han, Le~Sun, and
  Hua Wu. 2022.
\newblock \href {https://doi.org/10.18653/V1/2022.ACL-LONG.395} {Unified
  structure generation for universal information extraction}.
\newblock In \emph{Proceedings of the 60th Annual Meeting of the Association
  for Computational Linguistics (Volume 1: Long Papers), {ACL} 2022, Dublin,
  Ireland, May 22-27, 2022}, pages 5755--5772. Association for Computational
  Linguistics.

\bibitem[{Madani and Ez-zahout(2022)}]{madani2022review}
Rabie Madani and Abderrahmane Ez-zahout. 2022.
\newblock \href {https://doi.org/10.14569/IJACSA.2022.0130365} {A review-based
  context-aware recommender systems: Using custom ner and factorization
  machines}.
\newblock \emph{International Journal of Advanced Computer Science and
  Applications}, 13(3).

\bibitem[{Mo et~al.(2023)Mo, Tang, Liu, Wang, Xu, Wang, Wu, and Li}]{10094905}
Ying Mo, Hongyin Tang, Jiahao Liu, Qifan Wang, Zenglin Xu, Jingang Wang, Wei
  Wu, and Zhoujun Li. 2023.
\newblock \href {https://doi.org/10.1109/ICASSP49357.2023.10094905} {Multi-task
  transformer with relation-attention and type-attention for named entity
  recognition}.
\newblock In \emph{ICASSP 2023 - 2023 IEEE International Conference on
  Acoustics, Speech and Signal Processing (ICASSP)}, pages 1--5.

\bibitem[{Moll{\'a} et~al.(2006)Moll{\'a}, van Zaanen, and
  Smith}]{molla2006named}
Diego Moll{\'a}, Menno van Zaanen, and Daniel Smith. 2006.
\newblock \href {https://aclanthology.org/U06-1009} {Named entity recognition
  for question answering}.
\newblock In \emph{Proceedings of the Australasian Language Technology Workshop
  2006}, pages 51--58, Sydney, Australia.

\bibitem[{OpenAI(2023)}]{openai2023gpt4}
OpenAI. 2023.
\newblock \href {http://arxiv.org/abs/2303.08774} {Gpt-4 technical report}.

\bibitem[{Raffel et~al.(2020)Raffel, Shazeer, Roberts, Lee, Narang, Matena,
  Zhou, Li, and Liu}]{T5}
Colin Raffel, Noam Shazeer, Adam Roberts, Katherine Lee, Sharan Narang, Michael
  Matena, Yanqi Zhou, Wei Li, and Peter~J. Liu. 2020.
\newblock \href {http://jmlr.org/papers/v21/20-074.html} {Exploring the limits
  of transfer learning with a unified text-to-text transformer}.
\newblock \emph{J. Mach. Learn. Res.}, 21(1).

\bibitem[{Ratinov and Roth(2009)}]{ratinov2009design}
Lev Ratinov and Dan Roth. 2009.
\newblock \href {https://aclanthology.org/W09-1119} {Design challenges and
  misconceptions in named entity recognition}.
\newblock In \emph{Proceedings of the Thirteenth Conference on Computational
  Natural Language Learning ({C}o{NLL}-2009)}, pages 147--155, Boulder,
  Colorado. Association for Computational Linguistics.

\bibitem[{Sharma and Daniel~Jr(2019)}]{sharma2019bioflair}
Shreyas Sharma and Ron Daniel~Jr. 2019.
\newblock \href {https://arxiv.org/abs/1908.05760} {Bioflair: Pretrained pooled
  contextualized embeddings for biomedical sequence labeling tasks}.
\newblock \emph{arXiv preprint arXiv:1908.05760}.

\bibitem[{Shen et~al.(2023)Shen, Song, Tan, Li, Lu, and
  Zhuang}]{shen-etal-2023-diffusionner}
Yongliang Shen, Kaitao Song, Xu~Tan, Dongsheng Li, Weiming Lu, and Yueting
  Zhuang. 2023.
\newblock \href {https://doi.org/10.18653/v1/2023.acl-long.215}
  {{D}iffusion{NER}: Boundary diffusion for named entity recognition}.
\newblock In \emph{Proceedings of the 61st Annual Meeting of the Association
  for Computational Linguistics (Volume 1: Long Papers)}, pages 3875--3890,
  Toronto, Canada. Association for Computational Linguistics.

\bibitem[{Strakov{\'a} et~al.(2019)Strakov{\'a}, Straka, and
  Hajic}]{strakova2019neural}
Jana Strakov{\'a}, Milan Straka, and Jan Hajic. 2019.
\newblock \href {https://doi.org/10.18653/v1/P19-1527} {Neural architectures
  for nested {NER} through linearization}.
\newblock In \emph{Proceedings of the 57th Annual Meeting of the Association
  for Computational Linguistics}, pages 5326--5331, Florence, Italy.
  Association for Computational Linguistics.

\bibitem[{Strubell et~al.(2017)Strubell, Verga, Belanger, and
  McCallum}]{strubell-etal-2017-fast}
Emma Strubell, Patrick Verga, David Belanger, and Andrew McCallum. 2017.
\newblock \href {https://doi.org/10.18653/v1/D17-1283} {Fast and accurate
  entity recognition with iterated dilated convolutions}.
\newblock In \emph{Proceedings of the 2017 Conference on Empirical Methods in
  Natural Language Processing}, pages 2670--2680, Copenhagen, Denmark.
  Association for Computational Linguistics.

\bibitem[{Tjong Kim~Sang and De~Meulder(2003)}]{conll2003}
Erik~F. Tjong Kim~Sang and Fien De~Meulder. 2003.
\newblock \href {https://aclanthology.org/W03-0419} {Introduction to the
  {C}o{NLL}-2003 shared task: Language-independent named entity recognition}.
\newblock In \emph{Proceedings of the Seventh Conference on Natural Language
  Learning at {HLT}-{NAACL} 2003}, pages 142--147.

\bibitem[{Wadhwa et~al.(2023)Wadhwa, Amir, and
  Wallace}]{wadhwa-etal-2023-revisiting}
Somin Wadhwa, Silvio Amir, and Byron Wallace. 2023.
\newblock \href {https://doi.org/10.18653/v1/2023.acl-long.868} {Revisiting
  relation extraction in the era of large language models}.
\newblock In \emph{Proceedings of the 61st Annual Meeting of the Association
  for Computational Linguistics (Volume 1: Long Papers)}, pages 15566--15589,
  Toronto, Canada. Association for Computational Linguistics.

\bibitem[{Wang and Lu(2018)}]{Wang2018NeuralSH}
Bailin Wang and Wei Lu. 2018.
\newblock \href {https://api.semanticscholar.org/CorpusID:52916675} {Neural
  segmental hypergraphs for overlapping mention recognition}.
\newblock In \emph{Conference on Empirical Methods in Natural Language
  Processing}.

\bibitem[{Wang et~al.(2020)Wang, Shou, Chen, and Chen}]{wang2020pyramid}
Jue Wang, Lidan Shou, Ke~Chen, and Gang Chen. 2020.
\newblock \href {https://doi.org/10.18653/v1/2020.acl-main.525} {{P}yramid: A
  layered model for nested named entity recognition}.
\newblock In \emph{Proceedings of the 58th Annual Meeting of the Association
  for Computational Linguistics}, pages 5918--5928, Online. Association for
  Computational Linguistics.

\bibitem[{Wang et~al.(2022{\natexlab{a}})Wang, Li, Yan, Yan, Wang, Wu, and
  Xu}]{wang2022instructionner}
Liwen Wang, Rumei Li, Yang Yan, Yuanmeng Yan, Sirui Wang, Wei Wu, and Weiran
  Xu. 2022{\natexlab{a}}.
\newblock \href {http://arxiv.org/abs/2203.03903} {Instructionner: A multi-task
  instruction-based generative framework for few-shot ner}.

\bibitem[{Wang et~al.(2022{\natexlab{b}})Wang, Dou, Xiong, Zou, Zhang, Gui,
  Qiao, Cheng, and Huang}]{wang-etal-2022-miner}
Xiao Wang, Shihan Dou, Limao Xiong, Yicheng Zou, Qi~Zhang, Tao Gui, Liang Qiao,
  Zhanzhan Cheng, and Xuanjing Huang. 2022{\natexlab{b}}.
\newblock \href {https://doi.org/10.18653/v1/2022.acl-long.383} {{MINER}:
  Improving out-of-vocabulary named entity recognition from an information
  theoretic perspective}.
\newblock In \emph{Proceedings of the 60th Annual Meeting of the Association
  for Computational Linguistics (Volume 1: Long Papers)}, pages 5590--5600,
  Dublin, Ireland. Association for Computational Linguistics.

\bibitem[{Wang et~al.(2023)Wang, Zhou, Zu, Xia, Chen, Zhang, Zheng, Ye, Zhang,
  Gui et~al.}]{wang2023instructuie}
Xiao Wang, Weikang Zhou, Can Zu, Han Xia, Tianze Chen, Yuansen Zhang, Rui
  Zheng, Junjie Ye, Qi~Zhang, Tao Gui, et~al. 2023.
\newblock \href {https://arxiv.org/abs/2304.08085} {Instructuie: Multi-task
  instruction tuning for unified information extraction}.
\newblock \emph{arXiv preprint arXiv:2304.08085}.

\bibitem[{Wang et~al.(2019)Wang, Shang, Liu, Lu, Liu, and Han}]{conllpp}
Zihan Wang, Jingbo Shang, Liyuan Liu, Lihao Lu, Jiacheng Liu, and Jiawei Han.
  2019.
\newblock \href {https://doi.org/10.18653/v1/D19-1519} {{C}ross{W}eigh:
  Training named entity tagger from imperfect annotations}.
\newblock In \emph{Proceedings of the 2019 Conference on Empirical Methods in
  Natural Language Processing and the 9th International Joint Conference on
  Natural Language Processing (EMNLP-IJCNLP)}, pages 5154--5163, Hong Kong,
  China. Association for Computational Linguistics.

\bibitem[{Wei et~al.(2022{\natexlab{a}})Wei, Tay, Bommasani, Raffel, Zoph,
  Borgeaud, Yogatama, Bosma, Zhou, Metzler, Chi, Hashimoto, Vinyals, Liang,
  Dean, and Fedus}]{Emergent}
Jason Wei, Yi~Tay, Rishi Bommasani, Colin Raffel, Barret Zoph, Sebastian
  Borgeaud, Dani Yogatama, Maarten Bosma, Denny Zhou, Donald Metzler, Ed~H.
  Chi, Tatsunori Hashimoto, Oriol Vinyals, Percy Liang, Jeff Dean, and William
  Fedus. 2022{\natexlab{a}}.
\newblock Emergent abilities of large language models.
\newblock \emph{Trans. Mach. Learn. Res.}, 2022.

\bibitem[{Wei et~al.(2022{\natexlab{b}})Wei, Wang, Schuurmans, Bosma, Chi, Le,
  and Zhou}]{DBLP:journals/corr/abs-2201-11903}
Jason Wei, Xuezhi Wang, Dale Schuurmans, Maarten Bosma, Ed~H. Chi, Quoc Le, and
  Denny Zhou. 2022{\natexlab{b}}.
\newblock \href {http://arxiv.org/abs/2201.11903} {Chain of thought prompting
  elicits reasoning in large language models}.
\newblock \emph{CoRR}, abs/2201.11903.

\bibitem[{Wei et~al.(2022{\natexlab{c}})Wei, Wang, Schuurmans, Bosma, Ichter,
  Xia, Chi, Le, and Zhou}]{COT}
Jason Wei, Xuezhi Wang, Dale Schuurmans, Maarten Bosma, Brian Ichter, Fei Xia,
  Ed~H. Chi, Quoc~V. Le, and Denny Zhou. 2022{\natexlab{c}}.
\newblock Chain-of-thought prompting elicits reasoning in large language
  models.
\newblock In \emph{NeurIPS}.

\bibitem[{Xu et~al.(2017)Xu, Xu, Liang, Xie, Liang, Cui, and Xiao}]{CNDB}
Bo~Xu, Yong Xu, Jiaqing Liang, Chenhao Xie, Bin Liang, Wanyun Cui, and Yanghua
  Xiao. 2017.
\newblock \href {https://doi.org/https://doi.org/10.1007/978-3-319-60045-1_44}
  {Cn-dbpedia: {A} never-ending chinese knowledge extraction system}.
\newblock In \emph{Advances in Artificial Intelligence: From Theory to Practice
  - 30th International Conference on Industrial Engineering and Other
  Applications of Applied Intelligent Systems, {IEA/AIE} 2017, Arras, France,
  June 27-30, 2017, Proceedings, Part {II}}, volume 10351 of \emph{Lecture
  Notes in Computer Science}, pages 428--438. Springer.

\bibitem[{Yan et~al.(2021)Yan, Gui, Dai, Guo, Zhang, and Qiu}]{yan2021unified}
Hang Yan, Tao Gui, Junqi Dai, Qipeng Guo, Zheng Zhang, and Xipeng Qiu. 2021.
\newblock \href {https://doi.org/10.18653/v1/2021.acl-long.451} {A unified
  generative framework for various {NER} subtasks}.
\newblock In \emph{Proceedings of the 59th Annual Meeting of the Association
  for Computational Linguistics and the 11th International Joint Conference on
  Natural Language Processing (Volume 1: Long Papers)}, pages 5808--5822,
  Online. Association for Computational Linguistics.

\bibitem[{Yu et~al.(2020{\natexlab{a}})Yu, Bohnet, and
  Poesio}]{yu-etal-2020-named}
Juntao Yu, Bernd Bohnet, and Massimo Poesio. 2020{\natexlab{a}}.
\newblock \href {https://doi.org/10.18653/v1/2020.acl-main.577} {Named entity
  recognition as dependency parsing}.
\newblock In \emph{Proceedings of the 58th Annual Meeting of the Association
  for Computational Linguistics}, pages 6470--6476, Online. Association for
  Computational Linguistics.

\bibitem[{Yu et~al.(2020{\natexlab{b}})Yu, Bohnet, and Poesio}]{Yu2020NamedER}
Juntao Yu, Bernd Bohnet, and Massimo Poesio. 2020{\natexlab{b}}.
\newblock \href {https://api.semanticscholar.org/CorpusID:218630027} {Named
  entity recognition as dependency parsing}.
\newblock In \emph{Annual Meeting of the Association for Computational
  Linguistics}.

\end{thebibliography}

\section*{Language Resource References}
\label{lr:ref}
\bibliographystylelanguageresource{lrec-coling2024-natbib}
\bibliographylanguageresource{languageresource}


\end{document}